\documentclass{article} 
\usepackage{iclr2024_conference,times}
\usepackage{marvosym}

\usepackage{amsmath,amsfonts,bm}

\def\eqref#1{equation~\ref{#1}}

\def\1{\bm{1}}

\DeclareMathAlphabet{\mathsfit}{\encodingdefault}{\sfdefault}{m}{sl}
\SetMathAlphabet{\mathsfit}{bold}{\encodingdefault}{\sfdefault}{bx}{n}

\usepackage[colorlinks,
            linkcolor=magenta,
            anchorcolor=blue,
            citecolor=gray
            ]{hyperref}
\definecolor{emphypurple}{rgb}{0.302, 0.055, 0.659}
\usepackage{hyperref}
\usepackage{url}
\usepackage{hyperref}
\usepackage{xcolor}
\usepackage{pifont}
\usepackage{soul}
\usepackage{url}
\usepackage[T1]{fontenc}
\usepackage{siunitx}
\usepackage{graphicx}
\usepackage{float}
\usepackage{multirow}
\usepackage{color}
\usepackage{wrapfig}
\usepackage{booktabs}
\usepackage{amssymb}
\usepackage{subfigure}
\usepackage{listings}
\usepackage{makecell}
\usepackage{amssymb,mathrsfs,amsmath}
\usepackage{amsmath, bm}
\usepackage{colortbl}
\definecolor{xred}{HTML}{BD4242}
\definecolor{xblue}{HTML}{C7A085}
\definecolor{xblues}{HTML}{52B256}
\definecolor{xgreen}{HTML}{52B256}
\definecolor{xpurple}{HTML}{7F52B2}
\definecolor{xorange}{HTML}{FD9337}
\definecolor{xdotted}{HTML}{999999}
\definecolor{xgray}{HTML}{777777}
\definecolor{xcyan}{HTML}{80F5DC}
\definecolor{xpink}{HTML}{f690ea}
\definecolor{xgraycyan}{HTML}{82bceb}
\usepackage{tikz}
\usetikzlibrary{positioning, calc}
\usepackage[most]{tcolorbox}

\newcommand{\cmark}{\ding{51}}%
\newcommand{\xmark}{\ding{55}}%
\newcommand{\efficacyhigh}{\textcolor{green}{\cmark\cmark}}
\newcommand{\efficacymed}{\textcolor{orange}{\cmark}}
\newcommand{\efficacylow}{\textcolor{red}{\xmark}}
\tcbset{
	common/.style n args={2}{
		colframe={#1},
		colback={#1!5},
		colbacktitle={#1},
		lower separated=false,
		coltitle=white,
		boxrule=1pt,
		fonttitle=\bfseries,
		enhanced,
		breakable,
		top=8pt,
		before skip=8pt,
		attach boxed title to top left={
			yshift=-0.25cm,
			xshift=0.38cm,
		},
		boxed title style={
			boxrule=0pt,
			colframe=white,
			arc=5pt,
			outer arc=4pt,
		},
		separator sign={~~},
		overlay unbroken and last={
			\node[text=white, align=right, rounded corners, fill=#1, xshift=-4mm, minimum height=6mm, anchor=east] at (frame.south east) {#2};
		}
	},
	defstyle/.style={
		common={xpurple}{$\bm{\pi}$},
	},
	theoremstyle/.style={
		common={xgraycyan}{$\star$},
	},
	proofstyle/.style={
		common={xgreen}{\textbf{QED}},
	},
	examplestyle/.style={
		common={xblue}{$\bigstar$},
	},
    examplestyles/.style={
		common={xblues}{$\bigstar$},
	},
	notestyle/.style={
		common={xred}{\textbf{!}},
	},
	challangestyle/.style={
		common={xorange}{\textbf{?}},
	},
}

\newtcbtheorem{definition}{Definition}{defstyle}{def}
\newtcbtheorem{theorem}{Theorem}{theoremstyle}{theorem}
\newtcbtheorem{proof}{Proof}{proofstyle}{proof}
\newtcbtheorem{example}{Example}{examplestyle}{example}
\newtcbtheorem{examples}{Examples}{examplestyle}{example}
\newtcbtheorem{note}{Note}{notestyle}{note}
\newtcbtheorem{challange}{Challange}{challangestyle}{challange}

\usepackage[utf8]{inputenc} 
\usepackage[T1]{fontenc}    
\usepackage{hyperref}       
\usepackage{url}            
\usepackage{booktabs}       
\usepackage{amsfonts}       
\usepackage{nicefrac}       
\usepackage{microtype}      
\usepackage{xcolor}         

\title{Mastering Symbolic Operations: \\ Augmenting Language Models with Compiled Neural Networks}

\author{%
  Yixuan Weng $^{1*}$, Minjun Zhu$^{1,2}$\thanks{These authors contribute equally to this work. And {\Letter} means corresponding author.}, Fei Xia$^{1,2}$, Bin Li$^{3}$, Shizhu He$^{1,2, \textsuperscript{\Letter}}$ , Kang Liu$^{1,2, \textsuperscript{\Letter}}$, Jun Zhao$^{1,2}$\\
$^1$ The Laboratory of Cognition and Decision Intelligence for Complex Systems, IA, CAS \\
$^2$ School of Artificial Intelligence, University of Chinese Academy of Sciences\\
$^3$ College of Electrical and Information Engineering, Hunan University \\
\texttt{{wengsyx@gmail.com}, {\{shizhu.he, kliu, jzhao\}@nlpr.ia.ac.cn}}   \\
\url{https://github.com/wengsyx/Neural-Comprehension}  \\
}

\iclrfinalcopy 
\begin{document}

\maketitle

\begin{abstract}
    \vspace{-0.05cm}
Language models' (LMs) proficiency in handling deterministic symbolic reasoning and rule-based tasks remains limited due to their dependency implicit learning on textual data. To endow LMs with genuine rule comprehension abilities, we propose "Neural Comprehension" - a framework that synergistically integrates compiled neural networks (CoNNs) into the standard transformer architecture. CoNNs are neural modules designed to explicitly encode rules through artificially generated attention weights. By incorporating CoNN modules, the Neural Comprehension framework enables LMs to accurately and robustly execute rule-intensive symbolic tasks. Extensive experiments demonstrate the superiority of our approach over existing techniques in terms of length generalization, efficiency, and interpretability for symbolic operations. Furthermore, it can be applied to LMs across different model scales, outperforming tool-calling methods in arithmetic reasoning tasks while maintaining superior inference efficiency. Our work highlights the potential of seamlessly unifying explicit rule learning via CoNNs and implicit pattern learning in LMs, paving the way for true symbolic comprehension capabilities.
\end{abstract}

\section{Introduction}
    \vspace{-0.2cm}
\begin{wrapfigure}{t}{0.55\textwidth}
    \vspace{-0.85cm}
  \begin{center}
    \includegraphics[width=0.53\textwidth]{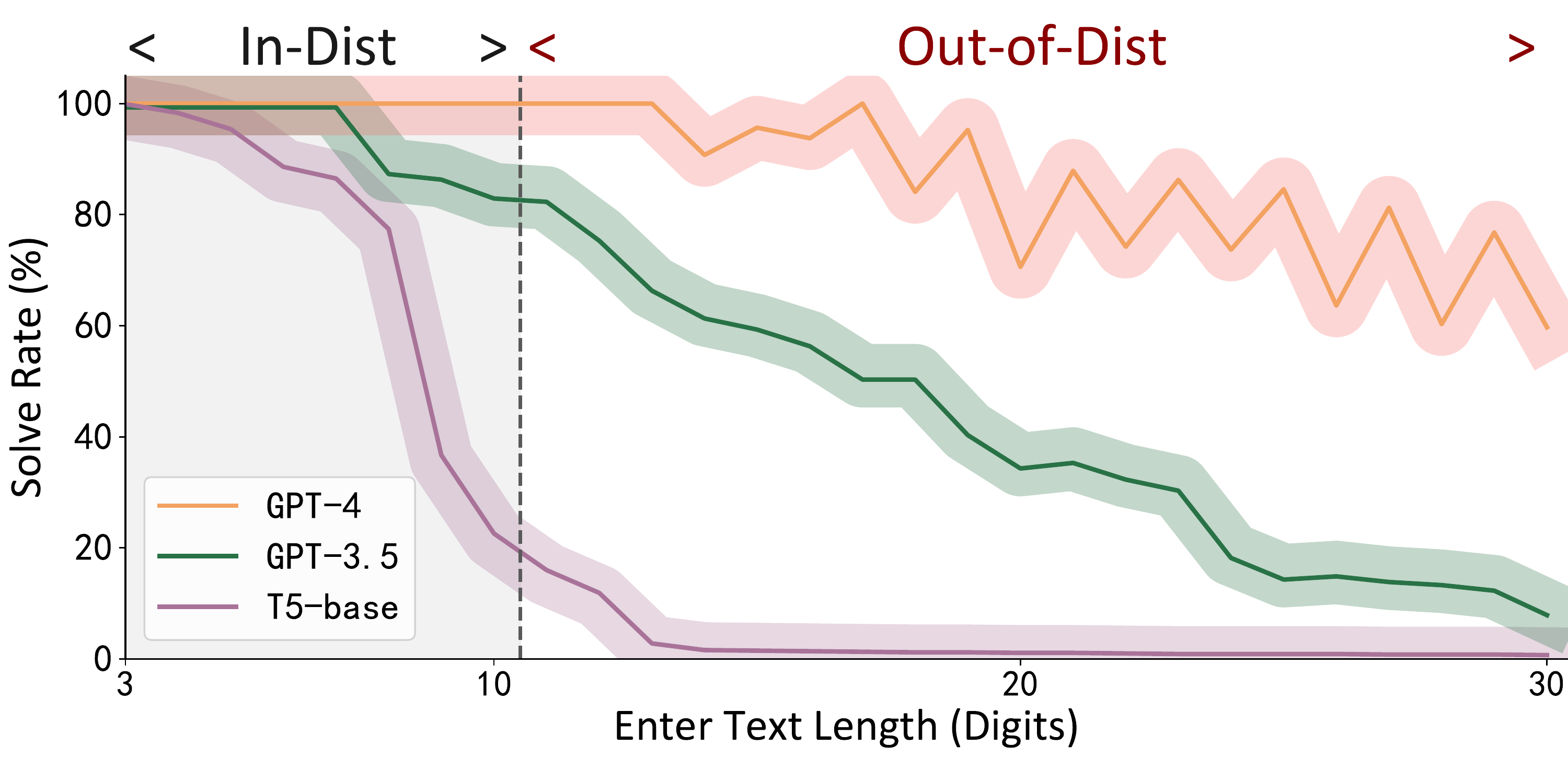}
          \vspace{-0.7cm}
  \end{center}
    \caption{The length generalization of T5 (with fine-tune), GPT-3.5 and GPT-4 (with few-shot) on symbolic operations (Addition) tasks.  To evaluate the model's proficiency, we conducted experiments on tasks ranging from 3 to 30 digits, with longer than 10 digits being out-of-distribution of training data.}
          \vspace{-0.25cm}
          \label{first}
\end{wrapfigure}

Language models (LMs), particularly large language models (LLMs), have exhibited impressive performance on complex reasoning tasks \citep{brown2020language,zhang2022opt,chowdhery2022palm,wei2022emergent,suzgun2022challenging}. Despite this, the proficiency of LMs in tackling deterministic symbolic reasoning and rule-based tasks is still limited \citep{welleckneural, razeghi2022impact}. For example, GPT-3's arithmetic performance declines with higher digit numbers \citep{brown2020language}, and its mathematical accuracy is influenced by word frequency in training data \citep{razeghi2022impact}. Moreover, length generalization \citep{anilexploring} remains a challenge even for 100-billion-parameter models, such as GPT-4 \citep{bubeck2023sparks}. We hypothesize that these limitations stem from LMs' dependency on implicitly learning rules from textual data. As illustrated in Figure \ref{first}, a simple length generalization experiment using addition tasks with varying numbers of digits highlights this limitation. Performance deteriorates as test length increases, indicating that these models strongly rely on statistical patterns in the data rather than capturing fundamental logical structures. This implicit learning of statistical patterns constrains LMs' accuracy in executing symbolic operations tasks.

We propose a transformer-based language model framework, termed "Neural Comprehension", which synergistically integrates a pretrained LM \citep{li2021pretrained} and compiled neural networks (CoNNs) \citep{weiss2021thinking}, combines their complementary strengths in a plug-and-play manner, to achieve high accuracy and robust performance. CoNNs are neural networks but the rules are explicitly coded through transformer-liked structures and attention. Therefore, the CoNN is human-controllable, executing rules through artificially generated attention weights, and can achieve perfect accuracy once compiled network is done. Neural Comprehension relys solely on neural networks without requiring additional tools. It employs a token-by-token generative method, analogous to GPT-3, where each token can be generated by either the pretrained LM or one of the CoNNs. The Neural Comprehension comprises a pretrained LM and multiple sets of CoNNs. The implementation of the Neural Comprehension framework facilitates the integration of rule-intensive abilities and reasoning capabilities into LMs, endowing them with genuine symbolic comprehension skills.

We conduct extensive experiments to evaluate the performance of our proposed Neural Comprehension method on a variety of rule-intensive tasks. Our experimental results demonstrate the effectiveness of our approach in comparison with existing state-of-the-art techniques, such as vanilla fine-tuning, few-shot learning, and Chain-of-Thought reasoning \citep{weichain}. Specifically, Neural Comprehension outperforms these methods in terms of accuracy, efficiency, and interpretability, showcasing its superiority in handling rule-intensive tasks. On the other hand, compared to the Tool-Based method \citep{mialon2023augmented}, Neural Comprehension provides a unified end-to-end neural network framework, eliminating the need for external interpreters and having higher inference efficiency. Historically, LMs are far from mastering robust symbolic task, such as symbolic operations and arithmetic reasoning \citep{stolfo-etal-2023-causal}. Our research provides a compelling case for LMs in neural network frameworks mastering symbolic operations, highlighting its potential to transform the landscape of symbolic reasoning and numerical computation capabilities for LMs.

\textbf{Contributions}\quad Our main contributions are as follows:

\begin{itemize}
     \item We pioneer the implementation of flawless execution rule-intensive symbolic operations for language models that rely on neural networks. By employing a plug-and-play way, we successfully integrate CoNNs, which are interpretable and human-controllable, into the language model. Our method facilitates direct rule deduction without the need for learning from conditional probabilities, leading to a more robust and effective approach. (\textbf{Section} \ref{section:method})
    \item We have built the AutoCoNN toolkit to make Neural Comprehension scalable, which leverages LLMs' contextual learning capabilities to automatically generate new CoNNs. Our method can be easily extended to various symbolic operations tasks. (\textbf{Seciton} \ref{section:toolkit})
\item Our experimental results on symbolic tasks and real-world arithmetic reasoning tasks demonstrate the superior performance of our method in comparison to existing techniques. Notably, our LM achieves flawless execution on symbolic reasoning tasks. (\textbf{Section} \ref{section:symbolic} \ref{section:symbolicreasoning})
\item It is worth noting that tool-based methods are only applicable to language models with code generation capabilities and require the cooperation of external interpreters. Our experiments demonstrate that the symbolic processing capabilities of neural understanding are on par with tool-based methods, but are applicable to models ranging from small ones like T5-Small (60M) to large ones like GLM-130B (130B) and GPT-3.5 (175B). (\textbf{Section} \ref{section:arithmetic})

\item We also studied the potential of combining multiple CoNNs and found that adding correlated CoNNs can continuously increase performance, while adding uncorrelated CoNNs rarely leads to performance degradation. This provides a new approach for model fusion, enabling the model to easily acquire new knowledge. (\textbf{Section} \ref{section:combination})

\end{itemize}

\section{Related Works}


\textbf{Pretrained Language Models} encompass those trained on general-purpose corpora \citep{lewis2019bart, scao2022bloom, MIR-2022-03-068} and specialized symbolic tasks \citep{geva2020injecting, lewkowycz2022solving,yang2023gpt}. They primarily aim to capture statistical patterns in language, which limits their capacity for symbolic reasoning. Symbolic reasoning involves manipulating abstract symbols and logical rules to derive new knowledge \citep{shindo2021neuro, yang2021learning} and necessitates the ability to extrapolate to novel situations and reason about concepts absent in the training data \citep{fujisawa2022logical}. Due to the constraints of gradient learning, LMs face challenges in wholly solving symbolic problems \citep{stolfo-etal-2023-causal}.

\textbf{In-Context Learning} has emerged as a promising approach to address these challenges \citep{dong2022survey} and closely approximate the predictors computed by gradient descent \citep{akyurek2022learning}. By prompting the language model to generate an explanation before generating an answer, the chain of thought \citep{weichain, TakeshiKojimaLargeLM, ZhuoshengZhang2022AutomaticCO, zhou2022least} encourages the model to think sequentially. This technique has been employed in various numerical and symbolic reasoning tasks, such as scratchpad prompting \citep{nye2021show} for length generalization \citep{anilexploring} and utilizing the chain of thought to perform arithmetic operations like summing pairs of single digits with carry \citep{zhou2022teaching}. However, this approach often necessitates substantial computational resources, and achieving perfect accuracy remains challenging.

\textbf{Augmented Language Models} have been proposed as an alternative, supplementing language models with external tools \citep{mialon2023augmented}. Examples include generating Python code for numerical reasoning \citep{gao2022pal, chen2022program} or incorporating tool usage as a pretraining task \citep{schick2023toolformer}. However, using external tools lacks a unified framework with language models and instead relies on the normativity of program generation.

\section{Preliminaries}

\begin{figure}[h]
\vspace{-0.35cm}
\begin{center}
	\includegraphics[width=0.88\textwidth]{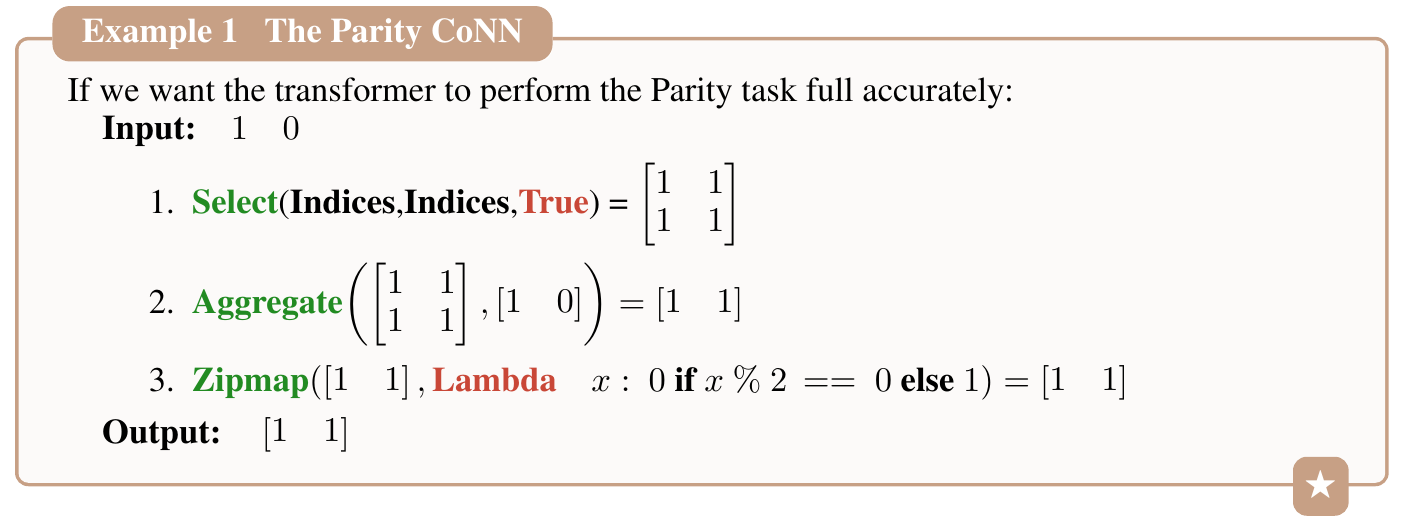}
\end{center}
\vspace{-0.8cm}
 \caption{Demonstration of the principles of \texttt{Parity} CoNN.}
 \vspace{-0.15cm}
 \label{fig:conn_example}
\end{figure}

\textbf{Compiled Neural Network (CoNN)}. CoNN is a transformer-based neural network leveraging artificially compiled attention weights to execute rules. CoNN has multiple attention layers and Multi-Layer Perceptron (MLP) layers, and each attention layer facilitates interactions between tokens. For example, in Figure \ref{fig:conn_example}, the multiplication of query and key elements representing a "{\textcolor[RGB]{34,139,34}{\textbf{Select}}}" operation in CoNN. Subsequent multiplication with value elements indicates an "{\textcolor[RGB]{34,139,34}{\textbf{Aggregate}}}" operation. The MLP layer is responsible for the token itself and is referred to as the "{\textcolor[RGB]{34,139,34}{\textbf{Zipmap}}}" operation \citep{weiss2021thinking}. By utilizing the three operations (Select, Aggregate, and Zipmap) to represent the sequence-to-sequence process, we can convert symbolic parity task into transformer weights \citep{lindner2023tracr}. CoNN can also stack multiple layers to address various human-defined rule problems, such as mathematical calculations and symbol operations \footnote{\textbf{Appendix} \ref{appendix:conn} provides a more detailed description of CoNN.}. Figure \ref{fig:conn_example} shows an example of \texttt{Parity} CoNN: The first step is to obtain a matrix of all ones to that of the sequence using the "{\textcolor[RGB]{34,139,34}{\textbf{Select}}}" operation. Secondly, the "{\textcolor[RGB]{34,139,34}{\textbf{Aggregate}}}" operation is used to combine the matrix obtained in the previous step with the input sequence (with the aim of calculating the total number of 0's and 1's in the sequence). The third step involves determining whether the total count is odd or even by "{\textcolor[RGB]{34,139,34}{\textbf{Zipmap}}}".

\section{Method}
 \vspace{-0.15cm}
Language models excel in language understanding tasks, but lack robust capabilities for symbolic tasks. We propose a Neural Comprehension framework that make CoNN guided by abstract rules into language models in a plug-and-play fashion, which integrates the language model's implicit learning parameters and CoNNs' explicit learning parameters. In Neural Comprehension, we designed CoNNs in neural networks with weights and structures directly encoding the symbolic rules  within the standard architecture of the LM to enhance deterministic rule comprehension and allow for deterministic execution of rule-based tasks.

\label{section:method}
\subsection{Neural Comprehension}

\begin{figure}[t]
\begin{center}
	\includegraphics[width=0.98\textwidth]{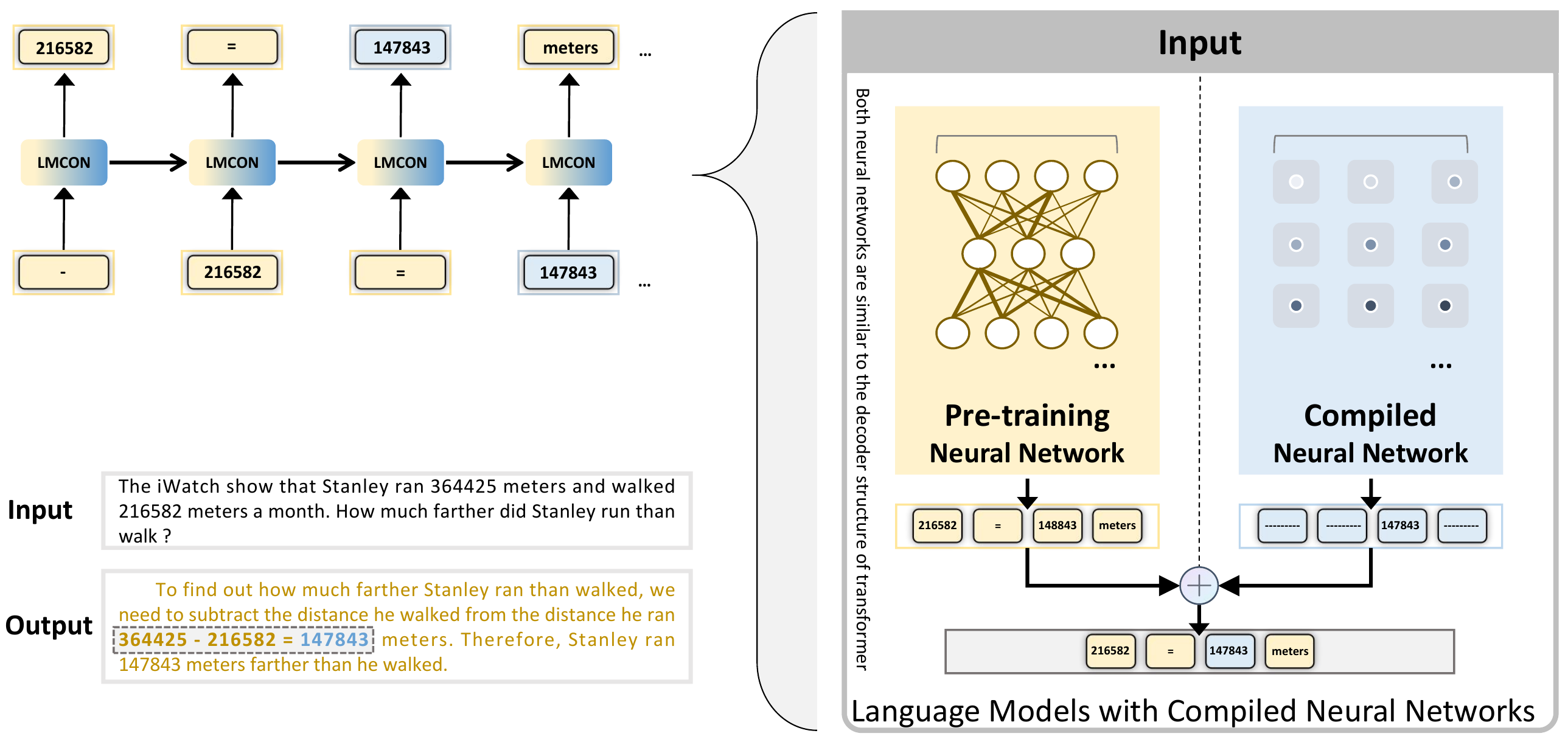}
\end{center}
\vspace{-0.1cm}
 \caption{The architecture of the proposed Neural Comprehension framework.}
\label{Neural_Comprehension}
\end{figure}

Neural Comprehension is a MoE-style \citep{DBLP:journals/corr/ShazeerMMDLHD17} neural network framework we designed, which is entirely composed of neural networks and maintains the generative seq2seq style. It uses predefined CoNNs as gating to determine when to output results from CoNNs. This approach is simple and plug-and-play, allowing combination with pretrained LMs. As illustrated in Figure~\ref{Neural_Comprehension}, the language model encodes the context and produces the textual and reasoning process context $D(x)$ step by step, a decoder architecture to generate the subsequent context step by step iteratively. And CoNNs handle sequence transformations involving rules, when a rule-required operation emerges, CoNN's attention is utilized to calculate specific values.  For example, in Figure~\ref{Neural_Comprehension}, when calculating \textit{364425-216582}, the only pretrained language model may output \textit{148843}, which is incorrect. However, the \texttt{Subtraction} CoNN can correct the result to \textit{147843} in the neural comprehension framework. This dynamically encoded process improves intermediate results interpretability and final result accuracy. Neural Comprehension combines LM and CoNNs in a piecewise function to perform gradient update. LM's hidden state output is $H_L=\left(H_{L_1} \cdots H_{L_{d_L}}\right)^{\top} \in \mathbb{R}^{d_L}, \quad H_{L_i} \in(0,1)$, and CoNN's output is $H_C=\left(H_{C_1} \cdots H_{C_{d_C}}\right)^{\top} \in \mathbb{R}^{d_C}, \quad H_{C_i} \in \{0,1\}$, 

\begin{equation}
\hat{i}=\underset{i}{\operatorname{argmax}}\left[\left(\begin{array}{c}
I_{d_L},0 \\
0,\beta I_{d_C}
\end{array}\right)\left(\begin{array}{c}
H_L, \\
H_C
\end{array}\right)\right],\quad \beta \in\{0,1\}
\label{equ:1}
\end{equation}

where $H_{C}$ is a one-hot vector, and $_{d_L}$ and $_{d_C}$ here refer to the vocabulary size of the Model's decode output. \footnote{It is worth noting that in this paper, for ease of implementation, the output vocabulary size of CoNNs' decode $_{d_C}$ is generally less than 100 due to limitations in computing resources (detailed information is shown in \textbf{Appendix Table 1}). } The Neural Comprehension combines the LLM's hidden state output, $H_L$, and CoNN's output, $H_C$, using identity matrices $I_{d_L}$ (for $d_L$) and $I_{d_C}$ (for $d_C$) to concatenate them for model fusion. Specifically, the hidden state representation matrix is obtained by extending the original hidden state representation matrix of the LM with the hidden state matrix on the CoNNs' vocabulary through a block matrix operation, resulting in a larger matrix.

\vspace{-0.15cm}

\subsection{Gradient Modification in Neural Comprehension}

To better appreciate the benefits of our method in handling rule-intensive tasks and improving accuracy, it is crucial to understand the gradient perspective of In-Context Learning (ICL). Recent studies on ICL algorithms have shown that the learning process of language models within the optimization process in ICL can be viewed as a search for suitable gradients to minimize the loss function. \citep{garg2022can,von2023transformers}. Due to the implicit learning nature of standard ICL methods, gradients learned from data may not always be ideal for addressing rule-intensive tasks. Therefore, our proposed method introduces an explicit learning component to provide more appropriate gradient updates for such tasks, ultimately leading to enhanced overall performance. We focus on elucidating the changes in the gradient introduced by the Neural Comprehension model, the gradient of the model during the execution of ICL can be partitioned into two categories based on the origin of the gradients:

\vspace{-0.2cm}
\begin{equation}
\text{Gradient} = \left\{\begin{array}{l}
G_{T} \quad \text{Text: Language Model}\\
G_{R} \quad \text{Rule: CoNNs}
\end{array}\right.
\vspace{-0.2cm}
\label{equ:gradient_partition}
\end{equation}

Here, $G_T$ represents the gradients derived implicitly from the language model (LM) and corresponds to the text-based learning aspect of the model. Conversely, $G_R$ represents the gradients explicitly derived from the CoNNs, encoding rule-based knowledge. {The specific computation process can be seen in Equation \ref{equ:1}. Note that the gradients' decomposition is only approximate and may not reflect the true generating process of text. The Neural Comprehension framework integrates both gradient sources to optimize the ICL process. In linear regression problems \citep{akyurek2022learning}, the loss function can be expressed as a piecewise function according to Equation \ref{equ:1}, here $P_1(x)$ is the LM and $P_2(x)$ is CoNN, the In-context-learner can be separate into two process:
\vspace{-0.12cm}
\begin{align}
    L&=\left\|y-\beta^{\top} x\right\|^2\\
     &= \left\{\begin{array}{l}
\left\|y-G_T^{\top} x\right\|^2 \quad x \in P_1(x)\\
\left\|y-G_R^{\top} x\right\|^2 \quad x \in P_2(x)
\end{array}\right.
\vspace{-0.56cm}
\label{equ:loss}
\end{align}

Based on the partitioned gradient as defined in Equation \ref{equ:gradient_partition}, the overall gradient of the Neural Comprehension model can be obtained by computing their individual gradients concerning the respective $\beta$:
\vspace{-0.3cm}
\begin{align}
\underbrace{\frac{\partial L}{\partial \beta}}_{\text{Gradient}} &= \left\{
\begin{array}{l}
\frac{\partial L}{\partial G_T} \quad x \in P_1(x)\\
\frac{\partial L}{\partial G_R} \quad x \in P_2(x)
\end{array}\right.
\vspace{-0.3cm}
\label{equ:par_gradient}
\end{align}

This partitioning allows the Neural Comprehension model to specifically address the gradient requirements of both implicit learning via LM and explicit learning via CoNNs. It is crucial to note that CoNNs are designed to minimize the loss associated with rule-based tasks, essentially providing an optimal gradient for tasks involving rule-intensive operations. This leads to a substantial improvement in the model's accuracy for rule-based tasks, as the gradient updates provided by CoNNs are more suitable for rule learning compared to the initially available gradients from the LM. By amalgamating both of gradient sources, the Neural Comprehension model achieves a more refined optimization of in-context learning. The Neural Comprehension model effectively balances the need for implicit and explicit learning within the ICL framework, leading to enhanced overall performance in terms of accuracy and interpretability.

\subsection{AutoCoNN Toolkit}
\label{section:toolkit}
To improve the scalability of Neural Comprehension frameworks, we propose the AutoCoNN toolkit, which can automatically generate new CoNNs and adapt Neural Comprehension given by "Instruct" and "Example", where "Instruct" serves as explicit symbolic definitions, and "Example" provides some input-output pairs for the operation. Considering that LLMs like GPT-4 can describe symbolic reasoning processes but cannot faithfully execute them \citep{cai2023large}, AutoCoNN Toolkit enables converting the reasoning process of symbols into CoNNs while maintaining a complete neural network framework without extra interpreters. The AutoCoNN process is divided into three steps. First, we provide 24 Tracr code \citep{lindner2023tracr} examples as context. Then the LLM is asked to generate 20 different Tracr codes by sampling decoding based on the "Instruct" and "Example". Finally, we convert these codes into pytorch-form \citep{NEURIPS2019_bdbca288} CoNNs and filter them on the pre-provided Examples to obtain accurate CoNNs. We provide further experimental analysis to test the performance of AutoCoNN in Appendix D. Meanwhile, the \texttt{Parity}, \texttt{Reverse}, \texttt{Copy} and \texttt{Last Letter} CoNN mentioned in the \textbf{Section} \ref{section:Experiment} are constructed by AutoCoNN, which demonstrates the practical value of AutoCoNN.

\section{Experiments and Result}
\label{section:Experiment}
\subsection{Symbolic Tasks}
\label{section:symbolic}
\begin{figure}[h]
 \vspace{-0.45cm}
\begin{center}
	\includegraphics[width=\textwidth]{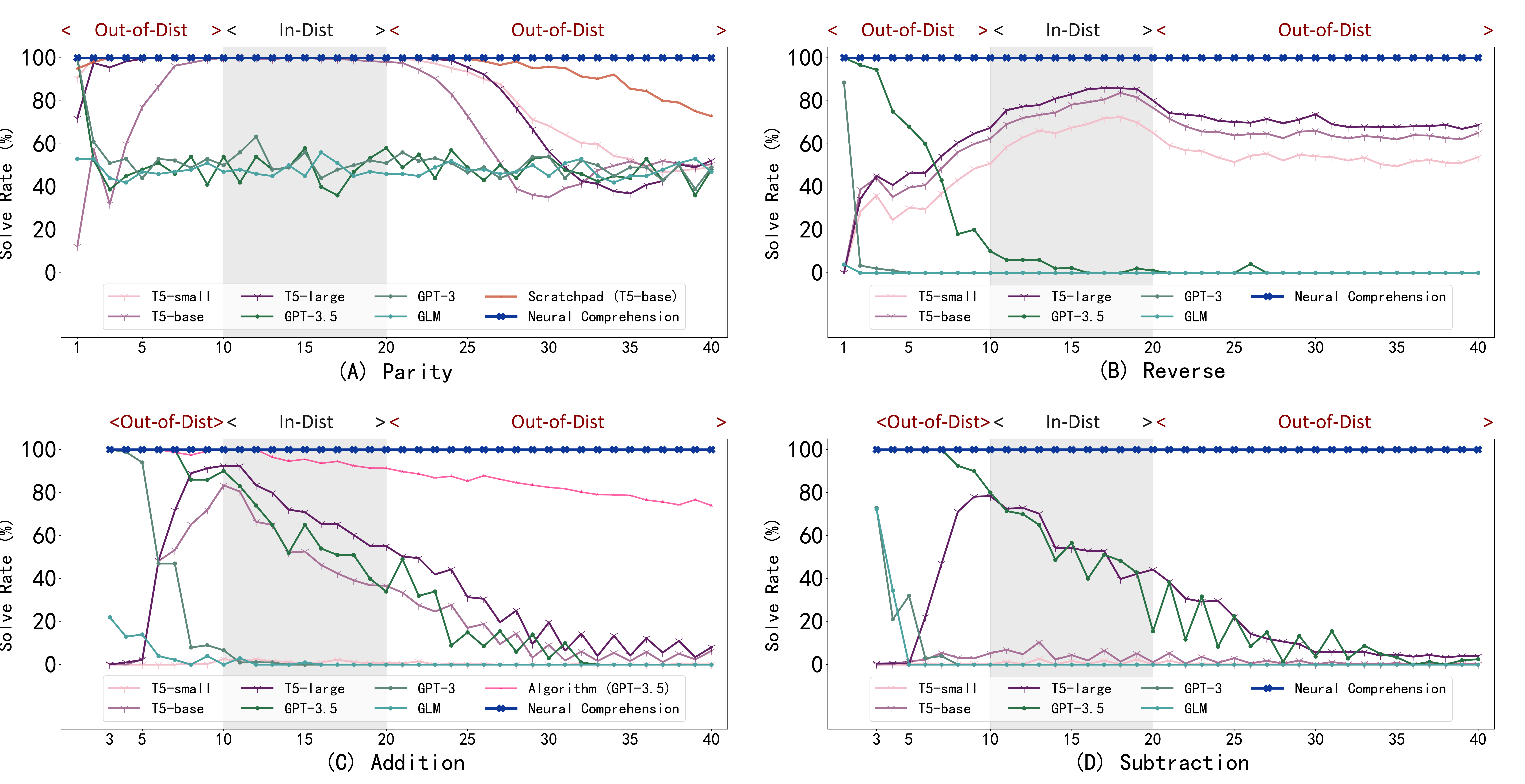}
 \vspace{-0.65cm}
 \label{fig:sybolic}
\end{center}
\caption{\small Comparison of Neural Comprehension and other implicit learning-based methods in symbolic operations tasks to test length generalization performance. In this, the T5 model uses the Vanilla Fine-tune method for learning, and LLMs use the Few-shot learning method. In Neural Comprehension, each task has a different CoNN, namely \texttt{Parity}, \texttt{Reverse}, \texttt{Addition}, and \texttt{Subtraction}.}
\label{figure:1}
\end{figure}
\begin{table}[h]
\resizebox{\textwidth}{!}{%
\begin{tabular}{l|cccc}
\midrule \midrule
\textbf{Techniques}  & \textbf{In-distribution} & \textbf{Out-of-distribution} & \textbf{Time and Space Complexity}& \textbf{Interpretability}  \\ \midrule
Vanilla Fine-tune (For LM)& \efficacyhigh & \efficacylow & \efficacyhigh &\efficacylow \\
Vanilla Few-shot (For LLM) & \efficacymed & \efficacymed  & \efficacyhigh&\efficacylow \\ 
Scratchpad \citep{anilexploring}& \efficacyhigh & \efficacymed & \efficacylow&\efficacymed \\
Algorithmic \citep{zhou2022teaching}& \efficacyhigh & \efficacymed & \efficacylow&\efficacymed \\
\textbf{Neural Comprehension (Ours)} & \efficacyhigh & \efficacyhigh  & \efficacyhigh& \efficacyhigh \\ 

\midrule \midrule
\end{tabular}}
\vspace{-0.2cm}
\caption{\small Performance on Symbolic operations tasks of five techniques that language models admit: (1) Vanilla Finetuning, (2) Vanilla Few-shot, (3) Scratchpad (Chain-of-Thought reasoning), (4) Algorithmic (Chain-of-Thought reasoning) and (5) Neural Comprehension. We find that the first four learning-based methods have different modes of failure regarding in and out-of-distribution coverage for symbolic operations. However, Neural Comprehension has strong advantages in terms of length generalization, efficiency, and interpretability. \efficacylow~signifies poor \efficacymed~signifies nontrivial, \efficacyhigh~signifies near-perfect performance. (*) Refers to task-dependency. }

\label{table:lg_options}
\end{table}

We conduct a length generalization experiment \citep{anilexploring} to examine the distinctions between the Neural Comprehension and learning-based methods, as depicted in Figure \ref{figure:1}. Our experimental design encompasses $1000 \times 40$ independent test sets, comprising problems with varying digit lengths from 1 to 40 digits. 10 to 20 digits within the range are provided by us for methods based on implicit learning for training; during the testing phase, this range is called In-Dist. Furthermore, we present results for both Scratchpad \citep{anilexploring} and Algorithmic \citep{zhou2022teaching} approaches.

The results of our experiment demonstrate that the Vanilla Fine-tune (\textcolor[RGB]{169,115,153}{red lines}) method performs optimally on the in-domain (10-20 digit) training set, while its performance deteriorates for both more simplistic and more intricate. This finding suggests that the absence of relevant samples in the training set may cause gradient descent-based language models to underperform on both simpler and more complex tasks. As further discussed in the \textbf{appendix D.1}, this phenomenon can be attributed to the inherent generalization limitations of statistical models and the position bias of language models.

Considering the Vanilla Few-shot method (\textcolor[RGB]{40,114,70}{green lines}), we determine that its performance is less impacted by the prompt sample range compared to Vanilla Fine-tune. Large language models, which are trained on extensive text corpora, excel at solving more straightforward problems such as symbolic operations within a ten-digit range. Nevertheless, performance remains below par for test sets with more than ten digits, even when prompted with 10-20 digit samples.

Observing CoT-like methods (we use GPT-3.5), including Scratchpad and Algorithmic, unveils their robust length generalization capabilities. Scratchpad works by requiring large language models to record intermediate steps, while Algorithmic employs a similar approach to record the carry operations involved in the addition process. This can be primarily attributed to their proficiency in decomposing complex problems into smaller incremental steps and maintaining intermediate states. However, these methods necessitate substantial computational resources, and extending the length beyond the input limit of the model becomes challenging.

Our study reveals that Neural Comprehension attains remarkably high accuracy in symbolic operations. This implies that Neural Comprehension, unlike conventional methods, does not rely on training data and remains unaffected by discrepancies in input lengths for in-distribution and out-of-distribution data. Consequently, it alleviates the requirement for step-by-step work tracking, and language models with CoNNs only need relatively fewer computational steps to execute sequence operations directly. Encoding rules into neural network modules endows us with greater interpretability, enabling language models to flawlessly perform purely symbolic operation tasks.

\subsection{Symbolic Reasoning}

\label{section:symbolicreasoning}
\begin{wrapfigure}{r}{0.618\textwidth}
  \begin{center}
  \vspace{-0.75cm}
    \includegraphics[width=0.57\textwidth]{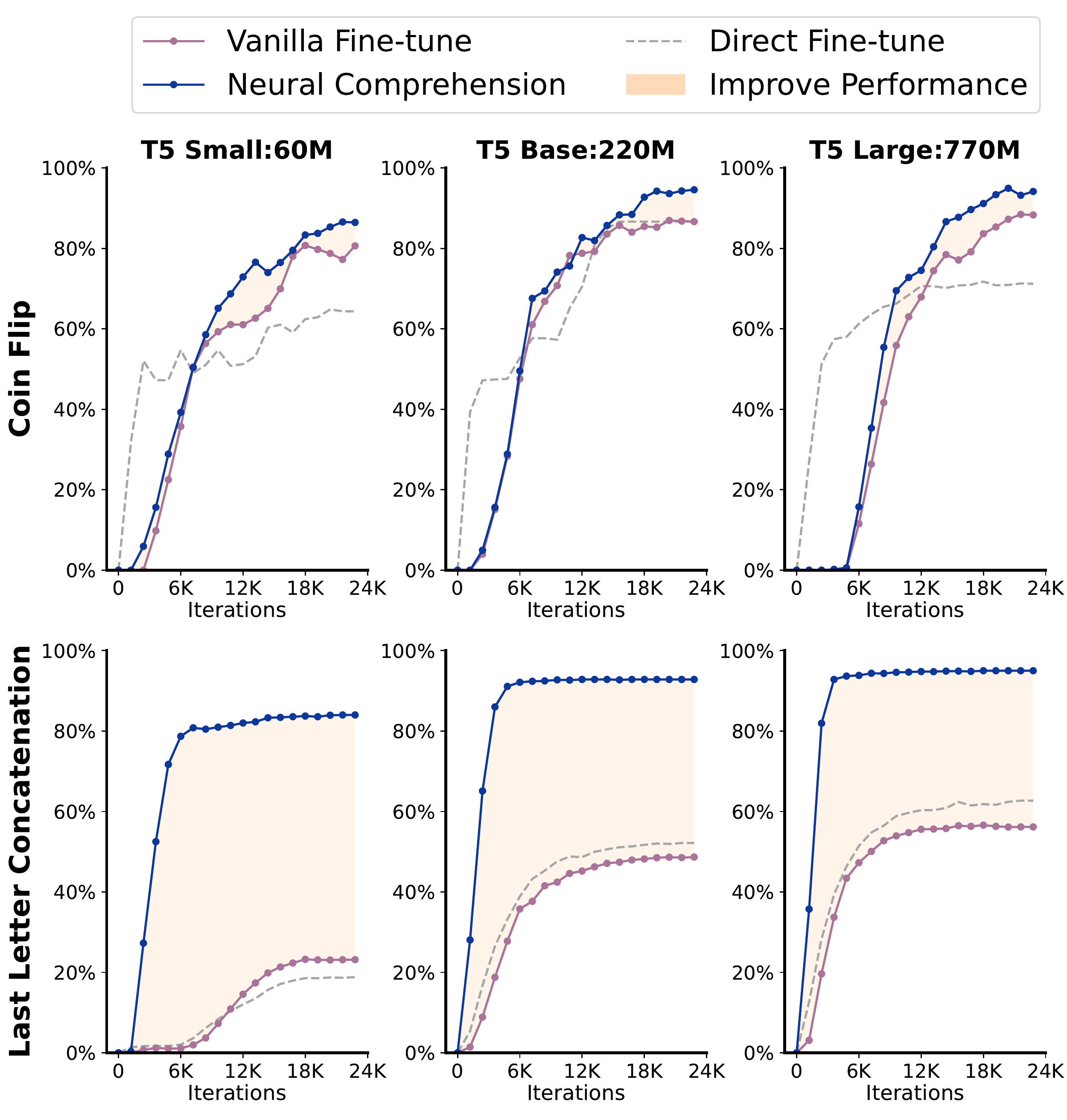}
  \end{center}
    \vspace{-0.05cm}
  \caption{In the iterative process of gradient descent during training. The {\textcolor[RGB]{14,55,154}{bleu line}} represents a language model that incorporates neural comprehension, and the \textcolor[RGB]{169,115,153}{red line} represents the original language model. Additionally, we provide \textcolor[RGB]{140,140,140}{Direct}, which is a direct prediction of the final result, as a reference.}
  \label{figure:2}
\vspace{-0.35cm}
\end{wrapfigure}

In this section, we investigate the performance of Neural Comprehension in terms of symbolic reasoning capabilities. Our hypothesis is that, although pretrained Language Models (LMs) demonstrate strong language understanding abilities, they lack the capacity to deduce and comprehend rules regarding symbolic reasoning tasks. Thus, we aim to evaluate whether the incorporation of compiled neural networks in the form of CoNNs can address this limitation and improve the LM's symbolic reasoning abilities.

To assess the performance of the rule comprehension component (CoNNs) in symbolic reasoning, we devise an experiment that measures the model's accuracy using intermediate processes and represents them in a "Chain of Thought"-like manner. In doing so, the experiment decomposes language understanding and rule comprehension explicitly into simpler outputs, avoiding the complexities of reasoning and additional error propagation in the models. Example outputs from this approach can be found in \textbf{Appendix F}. We observed that neural comprehension improves the symbolic reasoning capabilities of pretrained language models in most cases ({\textcolor[RGB]{14,55,154}{Neural Comprehension}} almost always outperforms {\textcolor[RGB]{169,115,153}{Vanilla Fine-tune}} in Figure \ref{figure:2}), and can fit faster. This observation suggests that the introduction of compiled neural networks has a positive impact on pretrained LMs, addressing rule comprehension limitations in symbolic reasoning tasks.

\subsection{Arithmetic Reasoning}
\label{section:arithmetic}
\vspace{-0.2cm}
\begin{figure}[h]
\begin{center}
	\includegraphics[width=\textwidth]{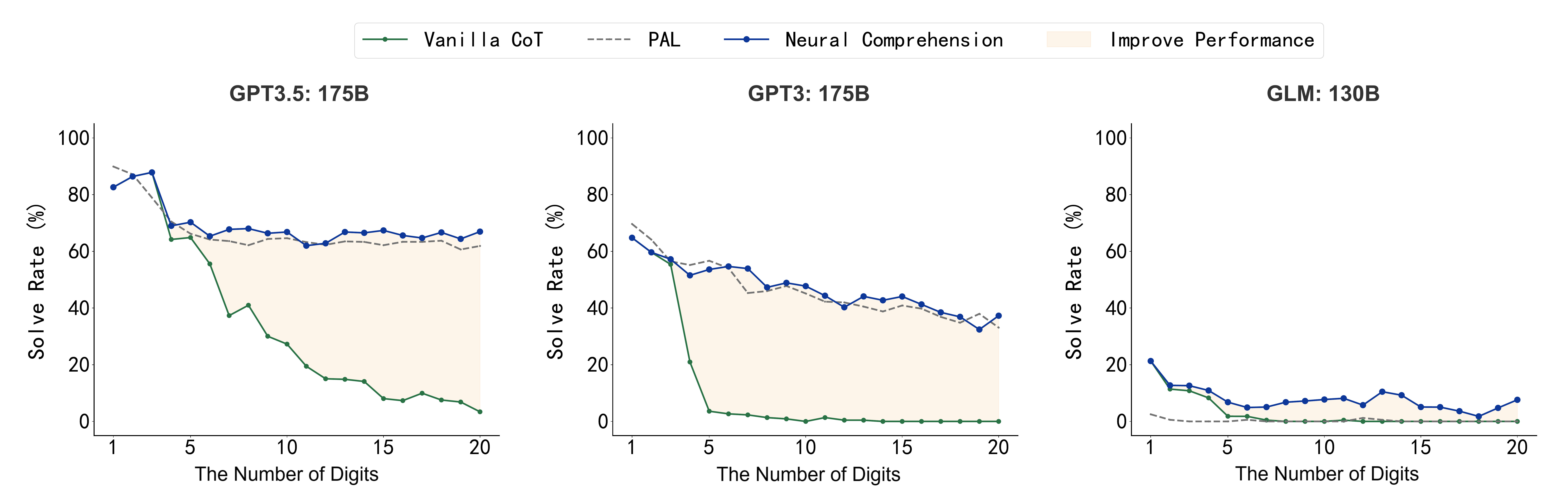}
\end{center}
\vspace{-0.35cm}
\caption{\small We conducted simulations of the AddSub dataset with varying digits by modifying the "lEquations" parameter. We then tested the performance of three LLMs with and without Neural Comprehension in generating CoT outputs for $\text{AddSub}^{+}$. And we reported the solve rates of three LLMs and compared the solve rates of using additional tools (PAL \citep{gao2022pal}).}
\vspace{-0.11cm}
\label{figure:3}
\end{figure}

\begin{table}[htbp]
\centering
\begin{minipage}[b]{0.315\linewidth}
\centering
\resizebox{\textwidth}{!}{%
\begin{tabular}{@{}lcccc@{}}
\toprule
\textbf{Addition} & \textbf{llama-2-7b} & \textbf{llama-2-70b} & \textbf{GLM-130B} & \textbf{Avg} \\
\midrule
CoT & 6.3 & 43.8 & 6.3 & 18.8 \\
PAL & \textbf{18.7} & 37.5 & 6.3 & 20.8 \\
NC (Ours) & 12.5 & \textbf{43.8} & \textbf{7.2} & \textbf{21.2} \\
\bottomrule
\end{tabular}%
}
\end{minipage}
\hfill
\begin{minipage}[b]{0.315\linewidth}
\centering
\resizebox{\textwidth}{!}{%
\begin{tabular}{@{}lcccc@{}}
\toprule
\textbf{Subtraction} & \textbf{llama-2-7b} & \textbf{llama-2-70b} & \textbf{GLM-130B} & \textbf{Avg} \\
\midrule
CoT & 6.3 & 27.9 & 1.8 & 12.0 \\
PAL & 7.2 & 31.5 & 3.6 & 14.1 \\
NC (Ours) & \textbf{9.9} & \textbf{32.4} & \textbf{4.5} & \textbf{15.6} \\
\bottomrule
\end{tabular}%
}
\end{minipage}
\hfill
\begin{minipage}[b]{0.315\linewidth}
\centering
\resizebox{\textwidth}{!}{%
\begin{tabular}{@{}lcccc@{}}
\toprule
\textbf{Mixed} & \textbf{llama-2-7b} & \textbf{llama-2-70b} & \textbf{GLM-130B} & \textbf{Avg} \\
\midrule
CoT & 11.1 & 16.7 & 0.0 & 9.3 \\
PAL & 11.1 & 27.8 & 5.6 & 14.8 \\
NC (Ours) & \textbf{27.8} & \textbf{33.3} & \textbf{5.6} & \textbf{22.2} \\
\bottomrule
\end{tabular}%
}
\end{minipage}

\caption{Experiments on the addition and subtraction subset for GSM8K-Hard dataset showing performance comparisons across different models and methods: Only Addition (left), Only Subtraction (center), and Mixed addition and subtraction (right), using Vanilla CoT, PAL, and NC (Neural Comprehension) methods.}
\label{tab:gsmhard}
\end{table}

Arithmetic reasoning serves as a suitable testbed for evaluating language models and their ability to address real-world problems. In this study, we examine the $\text{AddSub}^{+}$ dataset variants that involve different digit lengths, utilizing the \texttt{Addition} and \texttt{Subtraction} models from the CoNNs family. Notably, the capabilities of Neural Comprehension extend beyond these tasks, as CoNNs can also simulate calculators that support multiplication and division operations, and potentially perform linear algebra computations or even in-context learning algorithms that employ backpropagation \citep{giannou2023looped}.

To evaluate the impact of Neural Comprehension on arithmetic reasoning, we compare the output of vanilla CoT language models and those incorporating Neural Comprehension, using the vanilla CoT baseline as a reference. As demonstrated in Figure \ref{figure:3} and Table \ref{tab:gsmhard}, the vanilla CoT model struggles to extrapolate and solve arithmetic problems involving longer digit lengths. However, integrating Neural Comprehension significantly improves the performance of language models on such complex arithmetic tasks. Since we only incorporated the \texttt{Addition} and \texttt{Subtraction} CoNNs, we attribute the observed performance enhancement to the increased computational accuracy of the language model. For further evidence, we present additional experimental results on widely-used arithmetic reasoning datasets in \textbf{Appendix} \ref{appendix:arithmetic}, which reinforce the benefits of using Neural Comprehension over the vanilla CoT model.

In comparison to language models employing external tools like PAL \citep{gao2022pal}, our findings suggest that Neural Comprehension offers greater flexibility for LM. Firstly, by design, it minimizes the necessity for additional processing steps or external tools, leading to an efficient direct computational approach. This contrasts with Tool-based methods that often require additional programming and execution steps, increasing complexity and computational resources. Moreover, CoNN maintains end-to-end differentiability, crucial for models adapting to new data or tasks. In contrast, Tool-based methods are non-differentiable, limiting their adaptability in reinforcement learning settings or tasks with sparse delayed rewards \citep{chung2022scaling,LongOuyang2022TrainingLM}. Furthermore, CoNN's modularity enhances performance across various model scales, applicable regardless of a language model's ability to call functions, unlike tools only operable in large, additionally code-trained models. Thus, the Neural Comprehension framework's efficiency, unified end-to-end neural network architecture, and extensive applicability constitute its distinct advantages over the Tool-based approach, offering a robust and scalable solution for a multitude of linguistic and computational challenges.

\subsection{Ablation and Analyses: Module Combination for Neural Comprehension}
\label{section:combination}

\begin{figure}[h]
\begin{center}
	\hspace{0.45cm}\includegraphics[width=0.94\textwidth]{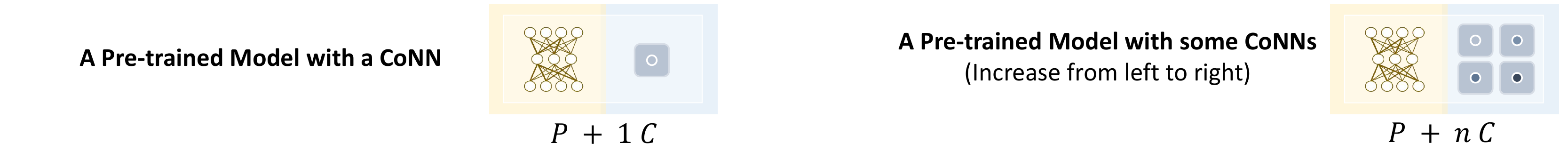}
 
 	\includegraphics[width=\textwidth]{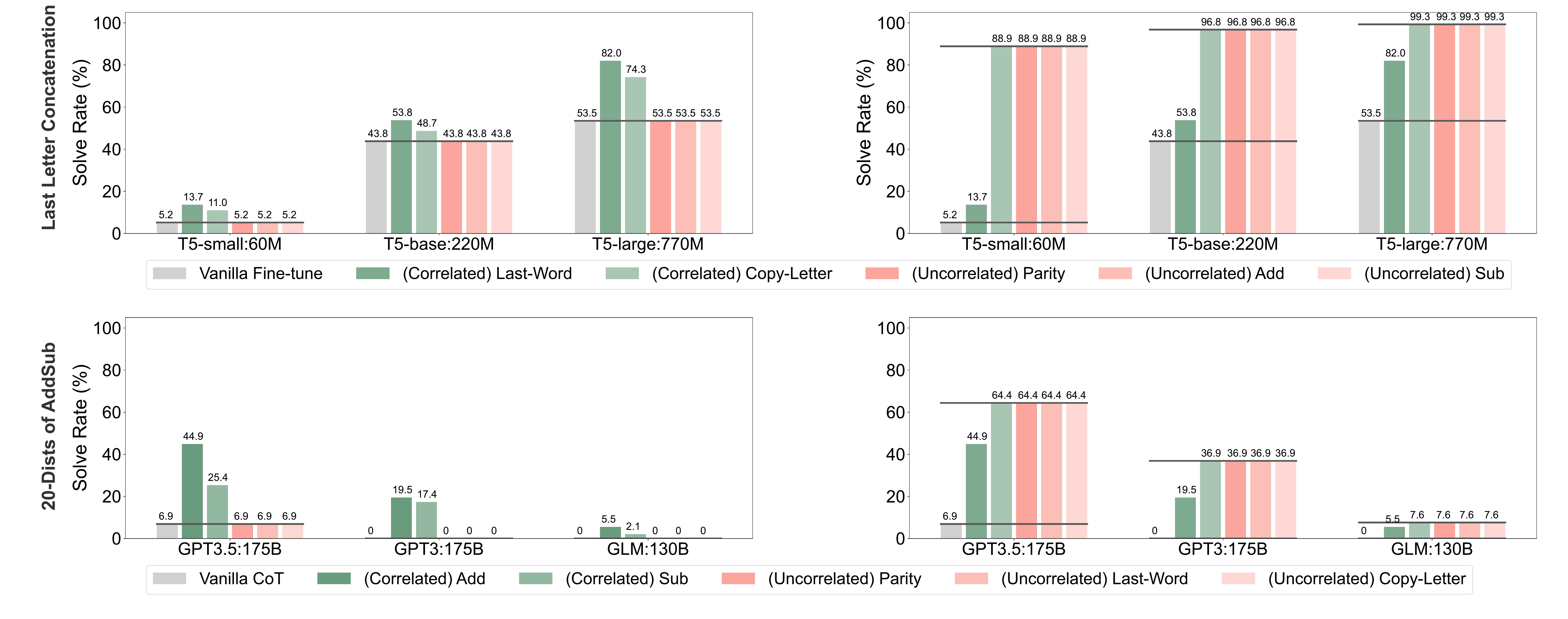}
\vspace{-0.4cm}

\end{center}

\caption{\small In Neural Comprehension framework, the performance of multiple different module combination is demonstrated. The left side shows the effect of combining a pretrained language model with a CoNN, while the right side shows the impact of combining a language model with multiple CoNNs. For different tasks, we categorize CoNNs as Correlated ({\textcolor[RGB]{104,156,125}{green}}) and Uncorrelated ({\textcolor[RGB]{252,166,156}{red}}), indicating whether the CoNN is related to the current task or not.}
\label{fig:two_model}
\end{figure}

Efficiently deploying multiple CoNNs is crucial for achieving exceptional Neural Comprehension performance. As depicted in Figure \ref{fig:two_model}, the amalgamation of distinct CoNNs, tailored for both symbolic and arithmetic reasoning tasks within the language model framework, can lead to remarkable benefits. Similar to ToolFormer \citep{schick2023toolformer}, we combine multiple different CoNNs into one framework, enabling the language model to have multiple capabilities. We conduct experiments on Last Letter Concatenation tass and $\text{AddSub}^{+}$ dataset, which shows the plug-and-play gating mechanism can still well control these CoNNs to output what should be output. It is observed that integrating pertinent CoNNs bolsters the performance of the initial language model, whereas the inclusion of unrelated language models rarely causes detrimental effects, regardless of whether single or multiple CoNNs are combined.

This can be ascribed to the refined design of the Neural Comprehension framework, which ensures the precise execution of assigned tasks by CoNNs without interference from irrelevant modules. Each CoNN module is adept at generating the appropriate output when needed, thereby preventing the emergence of erroneous results from unrelated components. Importantly, as seen in \textbf{Appendix B.3}, the parameter count for each CoNN module ranges from 1/1000 to 1/1000000 of that for GPT-3, and the experiments in \textbf{Appendix D.3} show that the inference latency in the neural understanding framework only increases by 1\%-3\% compared to Vanilla.

This observation underscores the remarkable scalability of the Neural Comprehension framework, which possesses the capability to not only accommodate existing knowledge concepts but also assimilate novel ones as the number of CoNNs expands. Theoretically, the integration of tens of thousands of CoNN modules within language models holds the potential to foster a comprehensive understanding of concepts.

\section{Conclusion}
We have observed that language models lack an intrinsic comprehension of rule-based concepts and explored how Neural Comprehension can integrate compiled neural networks into the language model framework in a simple and plug-and-play manner. On the one hand, we demonstrated the superiority of our approach over existing learning-based methods, where our method implements comparable improvements to external tools within the neural network framework but does not require additional interpreters. This also enables language models without coding capabilities to possess symbolic manipulation abilities. On the other hand, compared to external tools, gradients can propagate without proxies, allowing better integration and full differentiability. The Neural Comprehension solves the issue of language models themselves being unable to perform robust symbolic operations and providing a foundation for future work on unifying both implicit and explicit learning in language models and facilitating seamless integration.

\section*{Reproducibility Statement }
All CoNN models mentioned in this paper have been saved in Pytorch format in the \textbf{Supplementary Materials}, with dropout set to 0 to ensure deterministic outputs that conform to human-specified rules. The code for the AutoCoNN toolkit and Neural Comprehension framework in this paper can be found in the code portion of the \textbf{Supplementary Materials}. Details of all experiments setting referenced in this paper are included in \textbf{Appendix} \ref{appendix:model}. Detailed descriptions of all tasks, datasets, and baselines mentioned in this paper are provided in \textbf{Appendix} \ref{appendix:td}.  Details of all few-shot prompts referenced in this paper are included in \textbf{Appendix} \ref{appendix:example}.

\section*{Acknowledgements}
This work was supported by the National Key R\&D Program of China (No.2022ZD0118501) and the National Natural Science Foundation of China (No.U1936207, No.62376270,  No.62171183). Youth Innovation Promotion Association CAS, and OPPO Research Fund.

\subsection*{Acknowledgements for Colleagues}
We appreciate the interest shown in this work by our colleagues:
\begin{itemize}
    \item Haining Xie, Huanxuan Liao, Jiachun Li, Liang Gui, Pengfan Du, Pengfei Cao, Shaoru Guo, Wangtao Sun, Wenting Li, Xiusheng Huang, Yao Xu, Yifan Wei, Zhao Yang, Zhiqi Wang, Zhongtao Jiang, Zhuoran Jin, Ziyang Huang, who offered enthusiasm and backing.
\end{itemize}

\subsection*{Acknowledgements for friends}

We gratefully acknowledge the unwavering support and friendship from our community that served as the foundation for this research journey. Their companionship, through weekly board game and Werewolf gatherings, provided a reprieve from the rigors of research, allowing us to return refreshed and reinvigorated each week. Beyond the joyful times, they offered empathy during setbacks, perspective during challenges, and reassurance that progress awaits persistent effort. Their laughter, solidarity, and care kept us grounded, hopeful, and honest. The bonds formed enriched our lives immeasurably. We could not have navigated this meaningful journey without their understanding and encouragement. The impact of their compassion extends far beyond the technical contributions detailed herein. We express our sincerest appreciation, respect, and admiration for making this adventure a truly memorable experience:

\begin{itemize}
    \item Bingmei Sun, Boyuan Jiang, HaoChen Cao, Zixuan Cao, Donghui Li, Dongfang Suze, Ertan Zhuang, Jiajia Li, Mingwei Zhang, Lin Zhang, Mingwen Niu, Min Xiao, Qiaomu Tan, Tianyu Mu, Yuxin Liu, Xiaoyan Yu, Yuke Shi, Yixuan Li, Yang Zhou.
\end{itemize}

\bibliography{iclr2024_conference}

\begin{thebibliography}{71}
\providecommand{\natexlab}[1]{#1}
\providecommand{\url}[1]{\texttt{#1}}
\expandafter\ifx\csname urlstyle\endcsname\relax
  \providecommand{\doi}[1]{doi: #1}\else
  \providecommand{\doi}{doi: \begingroup \urlstyle{rm}\Url}\fi

\bibitem[Aky{\"u}rek et~al.(2022)Aky{\"u}rek, Schuurmans, Andreas, Ma, and Zhou]{akyurek2022learning}
Ekin Aky{\"u}rek, Dale Schuurmans, Jacob Andreas, Tengyu Ma, and Denny Zhou.
\newblock What learning algorithm is in-context learning? investigations with linear models.
\newblock \emph{arXiv preprint arXiv:2211.15661}, 2022.

\bibitem[Amini et~al.(2019)Amini, Gabriel, Lin, Koncel-Kedziorski, Choi, and Hajishirzi]{AidaAmini2019MathQATI}
Aida Amini, Saadia Gabriel, Peter Lin, Rik Koncel-Kedziorski, Yejin Choi, and Hannaneh Hajishirzi.
\newblock Mathqa: Towards interpretable math word problem solving with operation-based formalisms.
\newblock \emph{north american chapter of the association for computational linguistics}, 2019.

\bibitem[Anil et~al.(2022)Anil, Wu, Andreassen, Lewkowycz, Misra, Ramasesh, Slone, Gur-Ari, Dyer, and Neyshabur]{anilexploring}
Cem Anil, Yuhuai Wu, Anders~Johan Andreassen, Aitor Lewkowycz, Vedant Misra, Vinay~Venkatesh Ramasesh, Ambrose Slone, Guy Gur-Ari, Ethan Dyer, and Behnam Neyshabur.
\newblock Exploring length generalization in large language models.
\newblock In Alice~H. Oh, Alekh Agarwal, Danielle Belgrave, and Kyunghyun Cho (eds.), \emph{Advances in Neural Information Processing Systems}, 2022.
\newblock URL \url{https://openreview.net/forum?id=zSkYVeX7bC4}.

\bibitem[Arkil et~al.(2021)Arkil, Satwik, and Navin]{PatelArkil2021AreNM}
Patel Arkil, Bhattamishra Satwik, and Goyal Navin.
\newblock Are nlp models really able to solve simple math word problems?
\newblock 2021.

\bibitem[Brown et~al.(2020)Brown, Mann, Ryder, Subbiah, Kaplan, Dhariwal, Neelakantan, Shyam, Sastry, Askell, et~al.]{brown2020language}
Tom Brown, Benjamin Mann, Nick Ryder, Melanie Subbiah, Jared~D Kaplan, Prafulla Dhariwal, Arvind Neelakantan, Pranav Shyam, Girish Sastry, Amanda Askell, et~al.
\newblock Language models are few-shot learners.
\newblock \emph{Advances in neural information processing systems}, 33:\penalty0 1877--1901, 2020.

\bibitem[Bubeck et~al.(2023)Bubeck, Chandrasekaran, Eldan, Gehrke, Horvitz, Kamar, Lee, Lee, Li, Lundberg, Nori, Palangi, Ribeiro, and Zhang]{bubeck2023sparks}
Sébastien Bubeck, Varun Chandrasekaran, Ronen Eldan, Johannes Gehrke, Eric Horvitz, Ece Kamar, Peter Lee, Yin~Tat Lee, Yuanzhi Li, Scott Lundberg, Harsha Nori, Hamid Palangi, Marco~Tulio Ribeiro, and Yi~Zhang.
\newblock Sparks of artificial general intelligence: Early experiments with gpt-4, 2023.

\bibitem[Cai et~al.(2023)Cai, Wang, Ma, Chen, and Zhou]{cai2023large}
Tianle Cai, Xuezhi Wang, Tengyu Ma, Xinyun Chen, and Denny Zhou.
\newblock Large language models as tool makers.
\newblock \emph{arXiv preprint arXiv:2305.17126}, 2023.

\bibitem[Chen et~al.(2022)Chen, Ma, Wang, and Cohen]{chen2022program}
Wenhu Chen, Xueguang Ma, Xinyi Wang, and William~W Cohen.
\newblock Program of thoughts prompting: Disentangling computation from reasoning for numerical reasoning tasks.
\newblock \emph{arXiv preprint arXiv:2211.12588}, 2022.

\bibitem[Chowdhery et~al.(2022)Chowdhery, Narang, Devlin, Bosma, Mishra, Roberts, Barham, Chung, Sutton, Gehrmann, et~al.]{chowdhery2022palm}
Aakanksha Chowdhery, Sharan Narang, Jacob Devlin, Maarten Bosma, Gaurav Mishra, Adam Roberts, Paul Barham, Hyung~Won Chung, Charles Sutton, Sebastian Gehrmann, et~al.
\newblock Palm: Scaling language modeling with pathways.
\newblock \emph{arXiv preprint arXiv:2204.02311}, 2022.

\bibitem[Chung et~al.(2022)Chung, Hou, Longpre, Zoph, Tay, Fedus, Li, Wang, Dehghani, Brahma, Webson, Gu, Dai, Suzgun, Chen, Chowdhery, Castro-Ros, Pellat, Robinson, Valter, Narang, Mishra, Yu, Zhao, Huang, Dai, Yu, Petrov, Chi, Dean, Devlin, Roberts, Zhou, Le, and Wei]{chung2022scaling}
Hyung~Won Chung, Le~Hou, Shayne Longpre, Barret Zoph, Yi~Tay, William Fedus, Yunxuan Li, Xuezhi Wang, Mostafa Dehghani, Siddhartha Brahma, Albert Webson, Shixiang~Shane Gu, Zhuyun Dai, Mirac Suzgun, Xinyun Chen, Aakanksha Chowdhery, Alex Castro-Ros, Marie Pellat, Kevin Robinson, Dasha Valter, Sharan Narang, Gaurav Mishra, Adams Yu, Vincent Zhao, Yanping Huang, Andrew Dai, Hongkun Yu, Slav Petrov, Ed~H. Chi, Jeff Dean, Jacob Devlin, Adam Roberts, Denny Zhou, Quoc~V. Le, and Jason Wei.
\newblock Scaling instruction-finetuned language models, 2022.

\bibitem[Clark et~al.(2020)Clark, Tafjord, and Richardson]{ijcai2020p537}
Peter Clark, Oyvind Tafjord, and Kyle Richardson.
\newblock Transformers as soft reasoners over language.
\newblock In Christian Bessiere (ed.), \emph{Proceedings of the Twenty-Ninth International Joint Conference on Artificial Intelligence, {IJCAI-20}}, pp.\  3882--3890. International Joint Conferences on Artificial Intelligence Organization, 7 2020.
\newblock \doi{10.24963/ijcai.2020/537}.
\newblock URL \url{https://doi.org/10.24963/ijcai.2020/537}.
\newblock Main track.

\bibitem[Cobbe et~al.(2021)Cobbe, Kosaraju, Bavarian, Hilton, Nakano, Hesse, and Schulman]{cobbe2021training}
Karl Cobbe, Vineet Kosaraju, Mohammad Bavarian, Jacob Hilton, Reiichiro Nakano, Christopher Hesse, and John Schulman.
\newblock Training verifiers to solve math word problems.
\newblock \emph{arXiv preprint arXiv:2110.14168}, 2021.

\bibitem[Daull et~al.(2023)Daull, Bellot, Bruno, Martin, and Murisasco]{daull2023complex}
Xavier Daull, Patrice Bellot, Emmanuel Bruno, Vincent Martin, and Elisabeth Murisasco.
\newblock Complex qa and language models hybrid architectures, survey.
\newblock \emph{arXiv preprint arXiv:2302.09051}, 2023.

\bibitem[Devlin et~al.(2018)Devlin, Chang, Lee, and Toutanova]{devlin2018bert}
Jacob Devlin, Ming-Wei Chang, Kenton Lee, and Kristina Toutanova.
\newblock Bert: Pre-training of deep bidirectional transformers for language understanding.
\newblock \emph{arXiv preprint arXiv:1810.04805}, 2018.

\bibitem[Dong et~al.(2022)Dong, Li, Dai, Zheng, Wu, Chang, Sun, Xu, and Sui]{dong2022survey}
Qingxiu Dong, Lei Li, Damai Dai, Ce~Zheng, Zhiyong Wu, Baobao Chang, Xu~Sun, Jingjing Xu, and Zhifang Sui.
\newblock A survey for in-context learning.
\newblock \emph{arXiv preprint arXiv:2301.00234}, 2022.

\bibitem[Fujisawa \& Kanai(2022)Fujisawa and Kanai]{fujisawa2022logical}
Ippei Fujisawa and Ryota Kanai.
\newblock Logical tasks for measuring extrapolation and rule comprehension.
\newblock \emph{arXiv preprint arXiv:2211.07727}, 2022.

\bibitem[Gao et~al.(2022)Gao, Madaan, Zhou, Alon, Liu, Yang, Callan, and Neubig]{gao2022pal}
Luyu Gao, Aman Madaan, Shuyan Zhou, Uri Alon, Pengfei Liu, Yiming Yang, Jamie Callan, and Graham Neubig.
\newblock Pal: Program-aided language models.
\newblock \emph{arXiv preprint arXiv:2211.10435}, 2022.

\bibitem[Garg et~al.(2022)Garg, Tsipras, Liang, and Valiant]{garg2022can}
Shivam Garg, Dimitris Tsipras, Percy~S Liang, and Gregory Valiant.
\newblock What can transformers learn in-context? a case study of simple function classes.
\newblock \emph{Advances in Neural Information Processing Systems}, 35:\penalty0 30583--30598, 2022.

\bibitem[Geva et~al.(2020)Geva, Gupta, and Berant]{geva2020injecting}
Mor Geva, Ankit Gupta, and Jonathan Berant.
\newblock Injecting numerical reasoning skills into language models.
\newblock \emph{arXiv preprint arXiv:2004.04487}, 2020.

\bibitem[Giannou et~al.(2023)Giannou, Rajput, yong Sohn, Lee, Lee, and Papailiopoulos]{giannou2023looped}
Angeliki Giannou, Shashank Rajput, Jy~yong Sohn, Kangwook Lee, Jason~D. Lee, and Dimitris Papailiopoulos.
\newblock Looped transformers as programmable computers, 2023.

\bibitem[Hosseini et~al.(2014)Hosseini, Hajishirzi, Etzioni, and Kushman]{MohammadJavadHosseini2014LearningTS}
Mohammad~Javad Hosseini, Hannaneh Hajishirzi, Oren Etzioni, and Nate Kushman.
\newblock Learning to solve arithmetic word problems with verb categorization.
\newblock \emph{empirical methods in natural language processing}, 2014.

\bibitem[Hu et~al.(2019)Hu, Peng, Huang, and Li]{MinghaoHu2019AMM}
Minghao Hu, Yuxing Peng, Zhen Huang, and Dongsheng Li.
\newblock A multi-type multi-span network for reading comprehension that requires discrete reasoning.
\newblock \emph{empirical methods in natural language processing}, 2019.

\bibitem[Huang et~al.(2022)Huang, Gu, Hou, Wu, Wang, Yu, and Han]{huang2022large}
Jiaxin Huang, Shixiang~Shane Gu, Le~Hou, Yuexin Wu, Xuezhi Wang, Hongkun Yu, and Jiawei Han.
\newblock Large language models can self-improve.
\newblock \emph{arXiv preprint arXiv:2210.11610}, 2022.

\bibitem[Kojima et~al.(2022)Kojima, Gu, Reid, Matsuo, and Iwasawa]{TakeshiKojimaLargeLM}
Takeshi Kojima, Shixiang~Shane Gu, Machel Reid, Yutaka Matsuo, and Yusuke Iwasawa.
\newblock Large language models are zero-shot reasoners.
\newblock In Alice~H. Oh, Alekh Agarwal, Danielle Belgrave, and Kyunghyun Cho (eds.), \emph{Advances in Neural Information Processing Systems}, 2022.
\newblock URL \url{https://openreview.net/forum?id=e2TBb5y0yFf}.

\bibitem[Koncel-Kedziorski et~al.(2015)Koncel-Kedziorski, Hajishirzi, Sabharwal, Etzioni, and Ang]{RikKoncelKedziorski2015ParsingAW}
Rik Koncel-Kedziorski, Hannaneh Hajishirzi, Ashish Sabharwal, Oren Etzioni, and Siena~Dumas Ang.
\newblock Parsing algebraic word problems into equations.
\newblock \emph{Transactions of the Association for Computational Linguistics}, 2015.

\bibitem[Lewis et~al.(2019)Lewis, Liu, Goyal, Ghazvininejad, Mohamed, Levy, Stoyanov, and Zettlemoyer]{lewis2019bart}
Mike Lewis, Yinhan Liu, Naman Goyal, Marjan Ghazvininejad, Abdelrahman Mohamed, Omer Levy, Ves Stoyanov, and Luke Zettlemoyer.
\newblock Bart: Denoising sequence-to-sequence pre-training for natural language generation, translation, and comprehension.
\newblock \emph{arXiv preprint arXiv:1910.13461}, 2019.

\bibitem[Lewkowycz et~al.(2022)Lewkowycz, Andreassen, Dohan, Dyer, Michalewski, Ramasesh, Slone, Anil, Schlag, Gutman-Solo, et~al.]{lewkowycz2022solving}
Aitor Lewkowycz, Anders Andreassen, David Dohan, Ethan Dyer, Henryk Michalewski, Vinay Ramasesh, Ambrose Slone, Cem Anil, Imanol Schlag, Theo Gutman-Solo, et~al.
\newblock Solving quantitative reasoning problems with language models.
\newblock \emph{arXiv preprint arXiv:2206.14858}, 2022.

\bibitem[Li et~al.(2021{\natexlab{a}})Li, Chen, Liu, Weng, Sun, Li, Bai, and Hu]{li2021more}
Bin Li, Encheng Chen, Hongru Liu, Yixuan Weng, Bin Sun, Shutao Li, Yongping Bai, and Meiling Hu.
\newblock More but correct: Generating diversified and entity-revised medical response.
\newblock \emph{arXiv preprint arXiv:2108.01266}, 2021{\natexlab{a}}.

\bibitem[Li et~al.(2023)Li, Weng, Sun, and Li]{li2023learning}
Bin Li, Yixuan Weng, Bin Sun, and Shutao Li.
\newblock Learning to locate visual answer in video corpus using question.
\newblock In \emph{ICASSP 2023-2023 IEEE International Conference on Acoustics, Speech and Signal Processing (ICASSP)}, pp.\  1--5. IEEE, 2023.

\bibitem[Li et~al.(2024)Li, Weng, Xia, and Deng]{li2024towards}
Bin Li, Yixuan Weng, Fei Xia, and Hanjun Deng.
\newblock Towards better chinese-centric neural machine translation for low-resource languages.
\newblock \emph{Computer Speech \& Language}, 84:\penalty0 101566, 2024.

\bibitem[Li et~al.(2021{\natexlab{b}})Li, Tang, Zhao, and Wen]{li2021pretrained}
Junyi Li, Tianyi Tang, Wayne~Xin Zhao, and Ji-Rong Wen.
\newblock Pretrained language models for text generation: A survey, 2021{\natexlab{b}}.

\bibitem[Li et~al.(2022)Li, Choi, Chung, Kushman, Schrittwieser, Leblond, Eccles, Keeling, Gimeno, Dal~Lago, et~al.]{li2022competition}
Yujia Li, David Choi, Junyoung Chung, Nate Kushman, Julian Schrittwieser, R{\'e}mi Leblond, Tom Eccles, James Keeling, Felix Gimeno, Agustin Dal~Lago, et~al.
\newblock Competition-level code generation with alphacode.
\newblock \emph{Science}, 378\penalty0 (6624):\penalty0 1092--1097, 2022.

\bibitem[Lindner et~al.(2023)Lindner, Kram{\'a}r, Rahtz, McGrath, and Mikulik]{lindner2023tracr}
David Lindner, J{\'a}nos Kram{\'a}r, Matthew Rahtz, Thomas McGrath, and Vladimir Mikulik.
\newblock Tracr: Compiled transformers as a laboratory for interpretability.
\newblock \emph{arXiv preprint arXiv:2301.05062}, 2023.

\bibitem[Mialon et~al.(2023)Mialon, Dess{\`\i}, Lomeli, Nalmpantis, Pasunuru, Raileanu, Rozi{\`e}re, Schick, Dwivedi-Yu, Celikyilmaz, et~al.]{mialon2023augmented}
Gr{\'e}goire Mialon, Roberto Dess{\`\i}, Maria Lomeli, Christoforos Nalmpantis, Ram Pasunuru, Roberta Raileanu, Baptiste Rozi{\`e}re, Timo Schick, Jane Dwivedi-Yu, Asli Celikyilmaz, et~al.
\newblock Augmented language models: a survey.
\newblock \emph{arXiv preprint arXiv:2302.07842}, 2023.

\bibitem[Nijkamp et~al.(2022)Nijkamp, Pang, Hayashi, Tu, Wang, Zhou, Savarese, and Xiong]{nijkamp2022codegen}
Erik Nijkamp, Bo~Pang, Hiroaki Hayashi, Lifu Tu, Huan Wang, Yingbo Zhou, Silvio Savarese, and Caiming Xiong.
\newblock Codegen: An open large language model for code with multi-turn program synthesis.
\newblock \emph{arXiv preprint arXiv:2203.13474}, 2022.

\bibitem[Nye et~al.(2021)Nye, Andreassen, Gur-Ari, Michalewski, Austin, Bieber, Dohan, Lewkowycz, Bosma, Luan, et~al.]{nye2021show}
Maxwell Nye, Anders~Johan Andreassen, Guy Gur-Ari, Henryk Michalewski, Jacob Austin, David Bieber, David Dohan, Aitor Lewkowycz, Maarten Bosma, David Luan, et~al.
\newblock Show your work: Scratchpads for intermediate computation with language models.
\newblock \emph{arXiv preprint arXiv:2112.00114}, 2021.

\bibitem[Ouyang et~al.(2022)Ouyang, Wu, Jiang, Almeida, Wainwright, Mishkin, Zhang, Agarwal, Slama, Ray, Schulman, Hilton, Kelton, Miller, Simens, Askell, Welinder, Christiano, Leike, and Lowe]{LongOuyang2022TrainingLM}
Long Ouyang, Jeff Wu, Xu~Jiang, Diogo Almeida, Carroll Wainwright, Pamela Mishkin, Chong Zhang, Sandhini Agarwal, Katarina Slama, Alex Ray, John Schulman, Jacob Hilton, Fraser Kelton, Luke Miller, Maddie Simens, Amanda Askell, Peter Welinder, Paul Christiano, Jan Leike, and Ryan Lowe.
\newblock Training language models to follow instructions with human feedback.
\newblock 2022.

\bibitem[Paszke et~al.(2019)Paszke, Gross, Massa, Lerer, Bradbury, Chanan, Killeen, Lin, Gimelshein, Antiga, Desmaison, Kopf, Yang, DeVito, Raison, Tejani, Chilamkurthy, Steiner, Fang, Bai, and Chintala]{NEURIPS2019_bdbca288}
Adam Paszke, Sam Gross, Francisco Massa, Adam Lerer, James Bradbury, Gregory Chanan, Trevor Killeen, Zeming Lin, Natalia Gimelshein, Luca Antiga, Alban Desmaison, Andreas Kopf, Edward Yang, Zachary DeVito, Martin Raison, Alykhan Tejani, Sasank Chilamkurthy, Benoit Steiner, Lu~Fang, Junjie Bai, and Soumith Chintala.
\newblock Pytorch: An imperative style, high-performance deep learning library.
\newblock In H.~Wallach, H.~Larochelle, A.~Beygelzimer, F.~d\textquotesingle Alch\'{e}-Buc, E.~Fox, and R.~Garnett (eds.), \emph{Advances in Neural Information Processing Systems}, volume~32. Curran Associates, Inc., 2019.
\newblock URL \url{https://proceedings.neurips.cc/paper_files/paper/2019/file/bdbca288fee7f92f2bfa9f7012727740-Paper.pdf}.

\bibitem[Perez et~al.(2021)Perez, Kiela, and Cho]{perez2021true}
Ethan Perez, Douwe Kiela, and Kyunghyun Cho.
\newblock True few-shot learning with language models.
\newblock \emph{Advances in neural information processing systems}, 34:\penalty0 11054--11070, 2021.

\bibitem[Pi et~al.(2022)Pi, Liu, Chen, Ziyadi, Lin, Gao, Fu, Lou, and Chen]{XinyuPi2022ReasoningLP}
Xinyu Pi, Qian Liu, Bei Chen, Morteza Ziyadi, Zeqi Lin, Yan Gao, Qiang Fu, Jian-Guang Lou, and Weizhu Chen.
\newblock Reasoning like program executors.
\newblock 2022.

\bibitem[Qian et~al.(2022)Qian, Wang, Li, Li, and Yan]{qian2022limitations}
Jing Qian, Hong Wang, Zekun Li, Shiyang Li, and Xifeng Yan.
\newblock Limitations of language models in arithmetic and symbolic induction.
\newblock \emph{arXiv preprint arXiv:2208.05051}, 2022.

\bibitem[Razeghi et~al.(2022)Razeghi, Logan~IV, Gardner, and Singh]{razeghi2022impact}
Yasaman Razeghi, Robert~L Logan~IV, Matt Gardner, and Sameer Singh.
\newblock Impact of pretraining term frequencies on few-shot reasoning.
\newblock \emph{arXiv preprint arXiv:2202.07206}, 2022.

\bibitem[Roy \& Roth(2016)Roy and Roth]{SubhroRoy2016SolvingGA}
Subhro Roy and Dan Roth.
\newblock Solving general arithmetic word problems.
\newblock \emph{arXiv: Computation and Language}, 2016.

\bibitem[Scao et~al.(2022)Scao, Fan, Akiki, Pavlick, Ili{\'c}, Hesslow, Castagn{\'e}, Luccioni, Yvon, Gall{\'e}, et~al.]{scao2022bloom}
Teven~Le Scao, Angela Fan, Christopher Akiki, Ellie Pavlick, Suzana Ili{\'c}, Daniel Hesslow, Roman Castagn{\'e}, Alexandra~Sasha Luccioni, Fran{\c{c}}ois Yvon, Matthias Gall{\'e}, et~al.
\newblock Bloom: A 176b-parameter open-access multilingual language model.
\newblock \emph{arXiv preprint arXiv:2211.05100}, 2022.

\bibitem[Schick et~al.(2023)Schick, Dwivedi-Yu, Dess{\`\i}, Raileanu, Lomeli, Zettlemoyer, Cancedda, and Scialom]{schick2023toolformer}
Timo Schick, Jane Dwivedi-Yu, Roberto Dess{\`\i}, Roberta Raileanu, Maria Lomeli, Luke Zettlemoyer, Nicola Cancedda, and Thomas Scialom.
\newblock Toolformer: Language models can teach themselves to use tools.
\newblock \emph{arXiv preprint arXiv:2302.04761}, 2023.

\bibitem[Shazeer \& Stern(2018)Shazeer and Stern]{shazeer2018adafactor}
Noam Shazeer and Mitchell Stern.
\newblock Adafactor: Adaptive learning rates with sublinear memory cost, 2018.

\bibitem[Shazeer et~al.(2017)Shazeer, Mirhoseini, Maziarz, Davis, Le, Hinton, and Dean]{DBLP:journals/corr/ShazeerMMDLHD17}
Noam Shazeer, Azalia Mirhoseini, Krzysztof Maziarz, Andy Davis, Quoc~V. Le, Geoffrey~E. Hinton, and Jeff Dean.
\newblock Outrageously large neural networks: The sparsely-gated mixture-of-experts layer.
\newblock \emph{CoRR}, abs/1701.06538, 2017.
\newblock URL \url{http://arxiv.org/abs/1701.06538}.

\bibitem[Shindo et~al.(2021)Shindo, Dhami, and Kersting]{shindo2021neuro}
Hikaru Shindo, Devendra~Singh Dhami, and Kristian Kersting.
\newblock Neuro-symbolic forward reasoning.
\newblock \emph{arXiv preprint arXiv:2110.09383}, 2021.

\bibitem[Srivastava et~al.(2022)Srivastava, Rastogi, Rao, Shoeb, Abid, Fisch, Brown, Santoro, Gupta, Garriga-Alonso, Kluska, Lewkowycz, Agarwal, Power, Ray, Warstadt, Kocurek, Safaya, Tazarv, Xiang, Parrish, Nie, Hussain, Askell, Dsouza, Slone, Rahane, Iyer, Andreassen, Madotto, Santilli, Stuhlm\"uller, Dai, La, Lampinen, Zou, Jiang, Chen, Vuong, Gupta, Gottardi, Norelli, Venkatesh, Gholamidavoodi, Tabassum, Menezes, Kirubarajan, Mullokandov, Sabharwal, Herrick, Efrat, Erdem, Karaka\c\{s\}, Roberts, Loe, Zoph, Bojanowski, \"Ozyurt, Hedayatnia, Neyshabur, Inden, Stein, Ekmekci, Lin, Howald, Diao, Dour, Stinson, Argueta, Ram\'irez, Singh, Rathkopf, Meng, Baral, Wu, Callison-Burch, Waites, Voigt, Manning, Potts, Ramirez, Rivera, Siro, Raffel, Ashcraft, Garbacea, Sileo, Garrette, Hendrycks, Kilman, Roth, Freeman, Khashabi, Levy, Gonz\'alez, Perszyk, Hernandez, Chen, Ippolito, Gilboa, Dohan, Drakard, Jurgens, Datta, Ganguli, Emelin, Kleyko, Yuret, Chen, Tam, Hupkes, Misra, Buzan, Mollo, Yang, Lee, Shutova, Cubuk,
  Segal, Hagerman, Barnes, Donoway, Pavlick, Rodola, Lam, Chu, Tang, Erdem, Chang, Chi, Dyer, Jerzak, Kim, Manyasi, Zheltonozhskii, Xia, Siar, Mart\'inez-Plumed, Happ\'e, Chollet, Rong, Mishra, Winata, de~Melo, Kruszewski, Parascandolo, Mariani, Wang, Jaimovitch-L\'opez, Betz, Gur-Ari, Galijasevic, Kim, Rashkin, Hajishirzi, Mehta, Bogar, Shevlin, Sch\"utze, Yakura, Zhang, Wong, Ng, Noble, Jumelet, Geissinger, Kernion, Hilton, Lee, Fisac, Simon, Koppel, Zheng, Zou, Koco\'n, Thompson, Kaplan, Radom, Sohl-Dickstein, Phang, Wei, Yosinski, Novikova, Bosscher, Marsh, Kim, Taal, Engel, Alabi, Xu, Song, Tang, Waweru, Burden, Miller, Balis, Berant, Frohberg, Rozen, Hernandez-Orallo, Boudeman, Jones, Tenenbaum, Rule, Chua, Kanclerz, Livescu, Krauth, Gopalakrishnan, Ignatyeva, Markert, Dhole, Gimpel, Omondi, Mathewson, Chiafullo, Shkaruta, Shridhar, McDonell, Richardson, Reynolds, Gao, Zhang, Dugan, Qin, Contreras-Ochando, Morency, Moschella, Lam, Noble, Schmidt, He, Col\'on, Metz, \c\{S\}enel, Bosma, Sap, ter Hoeve,
  Farooqi, Faruqui, Mazeika, Baturan, Marelli, Maru, Quintana, Tolkiehn, Giulianelli, Lewis, Potthast, Leavitt, Hagen, Schubert, Baitemirova, Arnaud, McElrath, Yee, Cohen, Gu, Ivanitskiy, Starritt, Strube, Sw\k\{e\}drowski, Bevilacqua, Yasunaga, Kale, Cain, Xu, Suzgun, Tiwari, Bansal, Aminnaseri, Geva, Gheini, T, Peng, Chi, Lee, Krakover, Cameron, Roberts, Doiron, Nangia, Deckers, Muennighoff, Keskar, Iyer, Constant, Fiedel, Wen, Zhang, Agha, Elbaghdadi, Levy, Evans, Casares, Doshi, Fung, Liang, Vicol, Alipoormolabashi, Liao, Liang, Chang, Eckersley, Htut, Hwang, Mi\{\l\}kowski, Patil, Pezeshkpour, Oli, Mei, Lyu, Chen, Banjade, Rudolph, Gabriel, Habacker, Delgado, Milli\`ere, Garg, Barnes, Saurous, Arakawa, Raymaekers, Frank, Sikand, Novak, Sitelew, LeBras, Liu, Jacobs, Zhang, Salakhutdinov, Chi, Lee, Stovall, Teehan, Yang, Singh, Mohammad, Anand, Dillavou, Shleifer, Wiseman, Gruetter, Bowman, Schoenholz, Han, Kwatra, Rous, Ghazarian, Ghosh, Casey, Bischoff, Gehrmann, Schuster, Sadeghi, Hamdan, Zhou,
  Srivastava, Shi, Singh, Asaadi, Gu, Pachchigar, Toshniwal, Upadhyay, Debnath, Shakeri, Thormeyer, Melzi, Reddy, Makini, Lee, Torene, Hatwar, Dehaene, Divic, Ermon, Biderman, Lin, Prasad, Piantadosi, Shieber, Misherghi, Kiritchenko, Mishra, Linzen, Schuster, Li, Yu, Ali, Hashimoto, Wu, Desbordes, Rothschild, Phan, Wang, Nkinyili, Schick, Kornev, Telleen-Lawton, Tunduny, Gerstenberg, Chang, Neeraj, Khot, Shultz, Shaham, Misra, Demberg, Nyamai, Raunak, Ramasesh, Prabhu, Padmakumar, Srikumar, Fedus, Saunders, Zhang, Vossen, Ren, Tong, Zhao, Wu, Shen, Yaghoobzadeh, Lakretz, Song, Bahri, Choi, Yang, Hao, Chen, Belinkov, Hou, Hou, Bai, Seid, Zhao, Wang, Wang, Wang, and Wu]{AarohiSrivastava2022BeyondTI}
Aarohi Srivastava, Abhinav Rastogi, Abhishek Rao, Abu Awal~Md Shoeb, Abubakar Abid, Adam Fisch, Adam~R. Brown, Adam Santoro, Aditya Gupta, Adri\`a Garriga-Alonso, Agnieszka Kluska, Aitor Lewkowycz, Akshat Agarwal, Alethea Power, Alex Ray, Alex Warstadt, Alexander~W. Kocurek, Ali Safaya, Ali Tazarv, Alice Xiang, Alicia Parrish, Allen Nie, Aman Hussain, Amanda Askell, Amanda Dsouza, Ambrose Slone, Ameet Rahane, Anantharaman~S. Iyer, Anders Andreassen, Andrea Madotto, Andrea Santilli, Andreas Stuhlm\"uller, Andrew Dai, Andrew La, Andrew Lampinen, Andy Zou, Angela Jiang, Angelica Chen, Anh Vuong, Animesh Gupta, Anna Gottardi, Antonio Norelli, Anu Venkatesh, Arash Gholamidavoodi, Arfa Tabassum, Arul Menezes, Arun Kirubarajan, Asher Mullokandov, Ashish Sabharwal, Austin Herrick, Avia Efrat, Aykut Erdem, Ayla Karaka\c\{s\}, B.~Ryan Roberts, Bao~Sheng Loe, Barret Zoph, Bart\{\l\}omiej Bojanowski, Batuhan \"Ozyurt, Behnam Hedayatnia, Behnam Neyshabur, Benjamin Inden, Benno Stein, Berk Ekmekci, Bill~Yuchen Lin, Blake
  Howald, Cameron Diao, Cameron Dour, Catherine Stinson, Cedrick Argueta, C\'esar~Ferri Ram\'irez, Chandan Singh, Charles Rathkopf, Chenlin Meng, Chitta Baral, Chiyu Wu, Chris Callison-Burch, Chris Waites, Christian Voigt, Christopher~D. Manning, Christopher Potts, Cindy Ramirez, Clara~E. Rivera, Clemencia Siro, Colin Raffel, Courtney Ashcraft, Cristina Garbacea, Damien Sileo, Dan Garrette, Dan Hendrycks, Dan Kilman, Dan Roth, Daniel Freeman, Daniel Khashabi, Daniel Levy, Daniel~Mosegu\'i Gonz\'alez, Danielle Perszyk, Danny Hernandez, Danqi Chen, Daphne Ippolito, Dar Gilboa, David Dohan, David Drakard, David Jurgens, Debajyoti Datta, Deep Ganguli, Denis Emelin, Denis Kleyko, Deniz Yuret, Derek Chen, Derek Tam, Dieuwke Hupkes, Diganta Misra, Dilyar Buzan, Dimitri~Coelho Mollo, Diyi Yang, Dong-Ho Lee, Ekaterina Shutova, Ekin~Dogus Cubuk, Elad Segal, Eleanor Hagerman, Elizabeth Barnes, Elizabeth Donoway, Ellie Pavlick, Emanuele Rodola, Emma Lam, Eric Chu, Eric Tang, Erkut Erdem, Ernie Chang, Ethan~A. Chi, Ethan
  Dyer, Ethan Jerzak, Ethan Kim, Eunice~Engefu Manyasi, Evgenii Zheltonozhskii, Fanyue Xia, Fatemeh Siar, Fernando Mart\'inez-Plumed, Francesca Happ\'e, Francois Chollet, Frieda Rong, Gaurav Mishra, Genta~Indra Winata, Gerard de~Melo, Germ\'an Kruszewski, Giambattista Parascandolo, Giorgio Mariani, Gloria Wang, Gonzalo Jaimovitch-L\'opez, Gregor Betz, Guy Gur-Ari, Hana Galijasevic, Hannah Kim, Hannah Rashkin, Hannaneh Hajishirzi, Harsh Mehta, Hayden Bogar, Henry Shevlin, Hinrich Sch\"utze, Hiromu Yakura, Hongming Zhang, Hugh~Mee Wong, Ian Ng, Isaac Noble, Jaap Jumelet, Jack Geissinger, Jackson Kernion, Jacob Hilton, Jaehoon Lee, Jaime~Fern\'andez Fisac, James~B. Simon, James Koppel, James Zheng, James Zou, Jan Koco\'n, Jana Thompson, Jared Kaplan, Jarema Radom, Jascha Sohl-Dickstein, Jason Phang, Jason Wei, Jason Yosinski, Jekaterina Novikova, Jelle Bosscher, Jennifer Marsh, Jeremy Kim, Jeroen Taal, Jesse Engel, Jesujoba Alabi, Jiacheng Xu, Jiaming Song, Jillian Tang, Joan Waweru, John Burden, John Miller,
  John~U. Balis, Jonathan Berant, J\"org Frohberg, Jos Rozen, Jose Hernandez-Orallo, Joseph Boudeman, Joseph Jones, Joshua~B. Tenenbaum, Joshua~S. Rule, Joyce Chua, Kamil Kanclerz, Karen Livescu, Karl Krauth, Karthik Gopalakrishnan, Katerina Ignatyeva, Katja Markert, Kaustubh~D. Dhole, Kevin Gimpel, Kevin Omondi, Kory Mathewson, Kristen Chiafullo, Ksenia Shkaruta, Kumar Shridhar, Kyle McDonell, Kyle Richardson, Laria Reynolds, Leo Gao, Li~Zhang, Liam Dugan, Lianhui Qin, Lidia Contreras-Ochando, Louis-Philippe Morency, Luca Moschella, Lucas Lam, Lucy Noble, Ludwig Schmidt, Luheng He, Luis~Oliveros Col\'on, Luke Metz, L\"utfi~Kerem \c\{S\}enel, Maarten Bosma, Maarten Sap, Maartje ter Hoeve, Maheen Farooqi, Manaal Faruqui, Mantas Mazeika, Marco Baturan, Marco Marelli, Marco Maru, Maria Jose~Ram\'irez Quintana, Marie Tolkiehn, Mario Giulianelli, Martha Lewis, Martin Potthast, Matthew~L. Leavitt, Matthias Hagen, M\'aty\'as Schubert, Medina~Orduna Baitemirova, Melody Arnaud, Melvin McElrath, Michael~A. Yee, Michael
  Cohen, Michael Gu, Michael Ivanitskiy, Michael Starritt, Michael Strube, Micha\{\l\} Sw\k\{e\}drowski, Michele Bevilacqua, Michihiro Yasunaga, Mihir Kale, Mike Cain, Mimee Xu, Mirac Suzgun, Mo~Tiwari, Mohit Bansal, Moin Aminnaseri, Mor Geva, Mozhdeh Gheini, Mukund~Varma T, Nanyun Peng, Nathan Chi, Nayeon Lee, Neta Gur-Ari Krakover, Nicholas Cameron, Nicholas Roberts, Nick Doiron, Nikita Nangia, Niklas Deckers, Niklas Muennighoff, Nitish~Shirish Keskar, Niveditha~S. Iyer, Noah Constant, Noah Fiedel, Nuan Wen, Oliver Zhang, Omar Agha, Omar Elbaghdadi, Omer Levy, Owain Evans, Pablo Antonio~Moreno Casares, Parth Doshi, Pascale Fung, Paul~Pu Liang, Paul Vicol, Pegah Alipoormolabashi, Peiyuan Liao, Percy Liang, Peter Chang, Peter Eckersley, Phu~Mon Htut, Pinyu Hwang, Piotr Mi\{\l\}kowski, Piyush Patil, Pouya Pezeshkpour, Priti Oli, Qiaozhu Mei, Qing Lyu, Qinlang Chen, Rabin Banjade, Rachel~Etta Rudolph, Raefer Gabriel, Rahel Habacker, Ram\'on~Risco Delgado, Rapha\"el Milli\`ere, Rhythm Garg, Richard Barnes,
  Rif~A. Saurous, Riku Arakawa, Robbe Raymaekers, Robert Frank, Rohan Sikand, Roman Novak, Roman Sitelew, Ronan LeBras, Rosanne Liu, Rowan Jacobs, Rui Zhang, Ruslan Salakhutdinov, Ryan Chi, Ryan Lee, Ryan Stovall, Ryan Teehan, Rylan Yang, Sahib Singh, Saif~M. Mohammad, Sajant Anand, Sam Dillavou, Sam Shleifer, Sam Wiseman, Samuel Gruetter, Samuel~R. Bowman, Samuel~S. Schoenholz, Sanghyun Han, Sanjeev Kwatra, Sarah~A. Rous, Sarik Ghazarian, Sayan Ghosh, Sean Casey, Sebastian Bischoff, Sebastian Gehrmann, Sebastian Schuster, Sepideh Sadeghi, Shadi Hamdan, Sharon Zhou, Shashank Srivastava, Sherry Shi, Shikhar Singh, Shima Asaadi, Shixiang~Shane Gu, Shubh Pachchigar, Shubham Toshniwal, Shyam Upadhyay, Shyamolima~(Shammie) Debnath, Siamak Shakeri, Simon Thormeyer, Simone Melzi, Siva Reddy, Sneha~Priscilla Makini, Soo-Hwan Lee, Spencer Torene, Sriharsha Hatwar, Stanislas Dehaene, Stefan Divic, Stefano Ermon, Stella Biderman, Stephanie Lin, Stephen Prasad, Steven~T. Piantadosi, Stuart~M. Shieber, Summer Misherghi,
  Svetlana Kiritchenko, Swaroop Mishra, Tal Linzen, Tal Schuster, Tao Li, Tao Yu, Tariq Ali, Tatsu Hashimoto, Te-Lin Wu, Th\'eo Desbordes, Theodore Rothschild, Thomas Phan, Tianle Wang, Tiberius Nkinyili, Timo Schick, Timofei Kornev, Timothy Telleen-Lawton, Titus Tunduny, Tobias Gerstenberg, Trenton Chang, Trishala Neeraj, Tushar Khot, Tyler Shultz, Uri Shaham, Vedant Misra, Vera Demberg, Victoria Nyamai, Vikas Raunak, Vinay Ramasesh, Vinay~Uday Prabhu, Vishakh Padmakumar, Vivek Srikumar, William Fedus, William Saunders, William Zhang, Wout Vossen, Xiang Ren, Xiaoyu Tong, Xinran Zhao, Xinyi Wu, Xudong Shen, Yadollah Yaghoobzadeh, Yair Lakretz, Yangqiu Song, Yasaman Bahri, Yejin Choi, Yichi Yang, Yiding Hao, Yifu Chen, Yonatan Belinkov, Yu~Hou, Yufang Hou, Yuntao Bai, Zachary Seid, Zhuoye Zhao, Zijian Wang, Zijie~J. Wang, Zirui Wang, and Ziyi Wu.
\newblock Beyond the imitation game: Quantifying and extrapolating the capabilities of language models.
\newblock 2022.

\bibitem[Stolfo et~al.(2023)Stolfo, Jin, Shridhar, Schoelkopf, and Sachan]{stolfo-etal-2023-causal}
Alessandro Stolfo, Zhijing Jin, Kumar Shridhar, Bernhard Schoelkopf, and Mrinmaya Sachan.
\newblock A causal framework to quantify the robustness of mathematical reasoning with language models.
\newblock In \emph{Proceedings of the 61st Annual Meeting of the Association for Computational Linguistics (Volume 1: Long Papers)}, pp.\  545--561, Toronto, Canada, July 2023. Association for Computational Linguistics.
\newblock \doi{10.18653/v1/2023.acl-long.32}.
\newblock URL \url{https://aclanthology.org/2023.acl-long.32}.

\bibitem[Sun et~al.(2022)Sun, Liu, Qiu, and Huang]{MIR-2022-03-068}
Tian-Xiang Sun, Xiang-Yang Liu, Xi-Peng Qiu, and  Xuan-Jing Huang.
\newblock Paradigm shift in natural language processing.
\newblock \emph{Machine Intelligence Research}, 19\penalty0 (3):\penalty0 169--183, 2022.
\newblock ISSN 2731-538X.
\newblock \doi{10.1007/s11633-022-1331-6}.
\newblock URL \url{https://www.mi-research.net/en/article/doi/10.1007/s11633-022-1331-6}.

\bibitem[Suzgun et~al.(2022)Suzgun, Scales, Sch{\"a}rli, Gehrmann, Tay, Chung, Chowdhery, Le, Chi, Zhou, et~al.]{suzgun2022challenging}
Mirac Suzgun, Nathan Scales, Nathanael Sch{\"a}rli, Sebastian Gehrmann, Yi~Tay, Hyung~Won Chung, Aakanksha Chowdhery, Quoc~V Le, Ed~H Chi, Denny Zhou, et~al.
\newblock Challenging big-bench tasks and whether chain-of-thought can solve them.
\newblock \emph{arXiv preprint arXiv:2210.09261}, 2022.

\bibitem[Von~Oswald et~al.(2023)Von~Oswald, Niklasson, Randazzo, Sacramento, Mordvintsev, Zhmoginov, and Vladymyrov]{von2023transformers}
Johannes Von~Oswald, Eyvind Niklasson, Ettore Randazzo, Jo{\~a}o Sacramento, Alexander Mordvintsev, Andrey Zhmoginov, and Max Vladymyrov.
\newblock Transformers learn in-context by gradient descent.
\newblock In \emph{International Conference on Machine Learning}, pp.\  35151--35174. PMLR, 2023.

\bibitem[Wang et~al.(2023{\natexlab{a}})Wang, Chen, Qian, Gao, Wei, Wang, Tian, and Gao]{MIR-2022-07-224}
Xiao Wang, Guangyao Chen, Guangwu Qian, Pengcheng Gao, Xiao-Yong Wei, Yaowei Wang, Yonghong Tian, and Wen Gao.
\newblock Large-scale multi-modal pre-trained models: A comprehensive survey.
\newblock \emph{Machine Intelligence Research}, 20\penalty0 (4):\penalty0 447--482, 2023{\natexlab{a}}.
\newblock ISSN 2731-538X.
\newblock \doi{10.1007/s11633-022-1410-8}.
\newblock URL \url{https://www.mi-research.net/en/article/doi/10.1007/s11633-022-1410-8}.

\bibitem[Wang et~al.(2023{\natexlab{b}})Wang, Wei, Schuurmans, Le, Chi, Narang, Chowdhery, and Zhou]{XuezhiWangSelfConsistencyIC}
Xuezhi Wang, Jason Wei, Dale Schuurmans, Quoc~V Le, Ed~H. Chi, Sharan Narang, Aakanksha Chowdhery, and Denny Zhou.
\newblock Self-consistency improves chain of thought reasoning in language models.
\newblock In \emph{The Eleventh International Conference on Learning Representations}, 2023{\natexlab{b}}.
\newblock URL \url{https://openreview.net/forum?id=1PL1NIMMrw}.

\bibitem[Wei et~al.(2022{\natexlab{a}})Wei, Tay, Bommasani, Raffel, Zoph, Borgeaud, Yogatama, Bosma, Zhou, Metzler, et~al.]{wei2022emergent}
Jason Wei, Yi~Tay, Rishi Bommasani, Colin Raffel, Barret Zoph, Sebastian Borgeaud, Dani Yogatama, Maarten Bosma, Denny Zhou, Donald Metzler, et~al.
\newblock Emergent abilities of large language models.
\newblock \emph{arXiv preprint arXiv:2206.07682}, 2022{\natexlab{a}}.

\bibitem[Wei et~al.(2022{\natexlab{b}})Wei, Wang, Schuurmans, Bosma, Xia, Chi, Le, Zhou, et~al.]{weichain}
Jason Wei, Xuezhi Wang, Dale Schuurmans, Maarten Bosma, Fei Xia, Ed~H Chi, Quoc~V Le, Denny Zhou, et~al.
\newblock Chain-of-thought prompting elicits reasoning in large language models.
\newblock In \emph{Advances in Neural Information Processing Systems}, 2022{\natexlab{b}}.

\bibitem[Weiss et~al.(2021)Weiss, Goldberg, and Yahav]{weiss2021thinking}
Gail Weiss, Yoav Goldberg, and Eran Yahav.
\newblock Thinking like transformers.
\newblock In \emph{International Conference on Machine Learning}, pp.\  11080--11090. PMLR, 2021.

\bibitem[Welleck et~al.()Welleck, Kulikov, Roller, Dinan, Cho, and Weston]{welleckneural}
Sean Welleck, Ilia Kulikov, Stephen Roller, Emily Dinan, Kyunghyun Cho, and Jason Weston.
\newblock Neural text generation with unlikelihood training.
\newblock In \emph{International Conference on Learning Representations}.

\bibitem[Weng \& Li(2023)Weng and Li]{weng2023visual}
Yixuan Weng and Bin Li.
\newblock Visual answer localization with cross-modal mutual knowledge transfer.
\newblock In \emph{ICASSP 2023-2023 IEEE International Conference on Acoustics, Speech and Signal Processing (ICASSP)}, pp.\  1--5. IEEE, 2023.

\bibitem[Weng et~al.(2023{\natexlab{a}})Weng, Wang, Liao, He, Liu, Liu, and Zhao]{weng2023lmtuner}
Yixuan Weng, Zhiqi Wang, Huanxuan Liao, Shizhu He, Shengping Liu, Kang Liu, and Jun Zhao.
\newblock Lmtuner: An user-friendly and highly-integrable training framework for fine-tuning large language models.
\newblock \emph{arXiv preprint arXiv:2308.10252}, 2023{\natexlab{a}}.

\bibitem[Weng et~al.(2023{\natexlab{b}})Weng, Zhu, Xia, Li, He, Liu, Sun, Liu, and Zhao]{weng2022large}
Yixuan Weng, Minjun Zhu, Fei Xia, Bin Li, Shizhu He, Shengping Liu, Bin Sun, Kang Liu, and Jun Zhao.
\newblock Large language models are better reasoners with self-verification.
\newblock In \emph{Findings of the Association for Computational Linguistics: EMNLP 2023}, pp.\  2550--2575, 2023{\natexlab{b}}.

\bibitem[Xia et~al.(2022)Xia, Li, Weng, He, Liu, Sun, Li, and Zhao]{xia2022medconqa}
Fei Xia, Bin Li, Yixuan Weng, Shizhu He, Kang Liu, Bin Sun, Shutao Li, and Jun Zhao.
\newblock Medconqa: Medical conversational question answering system based on knowledge graphs.
\newblock In \emph{Proceedings of the The 2022 Conference on Empirical Methods in Natural Language Processing: System Demonstrations}, pp.\  148--158, 2022.

\bibitem[Yang \& Deng(2021)Yang and Deng]{yang2021learning}
Kaiyu Yang and Jia Deng.
\newblock Learning symbolic rules for reasoning in quasi-natural language.
\newblock \emph{arXiv preprint arXiv:2111.12038}, 2021.

\bibitem[Yang et~al.(2023)Yang, Ding, Lv, Jiang, He, Guo, Bai, and Tang]{yang2023gpt}
Zhen Yang, Ming Ding, Qingsong Lv, Zhihuan Jiang, Zehai He, Yuyi Guo, Jinfeng Bai, and Jie Tang.
\newblock Gpt can solve mathematical problems without a calculator, 2023.

\bibitem[Zhang et~al.(2022)Zhang, Roller, Goyal, Artetxe, Chen, Chen, Dewan, Diab, Li, Lin, et~al.]{zhang2022opt}
Susan Zhang, Stephen Roller, Naman Goyal, Mikel Artetxe, Moya Chen, Shuohui Chen, Christopher Dewan, Mona Diab, Xian Li, Xi~Victoria Lin, et~al.
\newblock Opt: Open pre-trained transformer language models.
\newblock \emph{arXiv preprint arXiv:2205.01068}, 2022.

\bibitem[Zhang et~al.(2023)Zhang, Zhang, Li, and Smola]{ZhuoshengZhang2022AutomaticCO}
Zhuosheng Zhang, Aston Zhang, Mu~Li, and Alex Smola.
\newblock Automatic chain of thought prompting in large language models.
\newblock In \emph{The Eleventh International Conference on Learning Representations}, 2023.
\newblock URL \url{https://openreview.net/forum?id=5NTt8GFjUHkr}.

\bibitem[Zhao et~al.(2023)Zhao, Zhang, and Zong]{MIR-2022-09-288}
Yang Zhao, Jiajun Zhang, and Chengqing Zong.
\newblock Transformer: A general framework from machine translation to others.
\newblock \emph{Machine Intelligence Research}, 20\penalty0 (4):\penalty0 514--538, 2023.
\newblock ISSN 2731-538X.
\newblock \doi{10.1007/s11633-022-1393-5}.
\newblock URL \url{https://www.mi-research.net/en/article/doi/10.1007/s11633-022-1393-5}.

\bibitem[Zhou et~al.(2022{\natexlab{a}})Zhou, Sch{\"a}rli, Hou, Wei, Scales, Wang, Schuurmans, Bousquet, Le, and Chi]{zhou2022least}
Denny Zhou, Nathanael Sch{\"a}rli, Le~Hou, Jason Wei, Nathan Scales, Xuezhi Wang, Dale Schuurmans, Olivier Bousquet, Quoc Le, and Ed~Chi.
\newblock Least-to-most prompting enables complex reasoning in large language models.
\newblock \emph{arXiv preprint arXiv:2205.10625}, 2022{\natexlab{a}}.

\bibitem[Zhou et~al.(2022{\natexlab{b}})Zhou, Nova, Larochelle, Courville, Neyshabur, and Sedghi]{zhou2022teaching}
Hattie Zhou, Azade Nova, Hugo Larochelle, Aaron Courville, Behnam Neyshabur, and Hanie Sedghi.
\newblock Teaching algorithmic reasoning via in-context learning.
\newblock \emph{arXiv preprint arXiv:2211.09066}, 2022{\natexlab{b}}.

\bibitem[Zhu et~al.(2022)Zhu, Weng, He, Liu, and Zhao]{zhu2022reasonchainqa}
Minjun Zhu, Yixuan Weng, Shizhu He, Kang Liu, and Jun Zhao.
\newblock Reasonchainqa: Text-based complex question answering with explainable evidence chains.
\newblock \emph{arXiv preprint arXiv:2210.08763}, 2022.

\end{thebibliography}
\bibliographystyle{iclr2024_conference}

\appendix

\section{Discussion of Limitations and Future Work}
\label{appendix:limitation}

We have presented a novel framework that integrates Compiled Neural Networks (CoNNs) with existing language models to bolster their rule understanding abilities. Although our approach has shown promising performance improvements on symbolic and arithmetic reasoning tasks, there are several limitations and potential avenues for future research that warrant further exploration.

A significant limitation of our current framework lies in the more efficient and natural incorporation of CoNNs into language models. Currently, our method employs a sparse neural network that treats the pretrained language model and CoNNs as separate modules. A more desirable solution is to leverage a dense neural network, simultaneously utilizing the benefits of both components. Examining the large-scale applicability of CoNNs is a beneficial endeavor. Although our experiments have been conducted on a relatively small scale (up to five stacked CoNNs), the advancements and abilities language models may gain from larger-scale combinations of CoNNs remain unclear. Exploring the scalability of our method and the performance advantages of deploying more complex CoNN architectures in language models could provide valuable insights into their potential.

Another promising area of research is the inclusion of explicit knowledge into CoNNs. While the current implementation centers on encoding rules into CoNNs, future work could exploit techniques from knowledge graphs to compile explicit knowledge into language models. This may significantly enhance language models' interpretability and knowledge representation capabilities, potentially resulting in improved performance on an even broader range of tasks.

In conclusion, our work on enhancing language models' rule understanding capabilities through CoNN integration has yielded promising results, albeit with some limitations and remaining challenges. By addressing these areas and extending our approach, we believe that it can ultimately lead to the development of more powerful, interpretable, and knowledge-rich language models.





\section{Compiled Neural Networks}
\label{appendix:conn}

In this section, we will discuss the concept, implementation, and potential of Compiled Neural Networks (CoNN), a type of neural network inspired from previous works on transformers. CoNNs can perform diverse tasks such as computer arithmetic and linear algebra, demonstrating a wide range of applications in LMs and beyond.

\subsection{Introduction}

Transformers have garnered significant attention due to their ability to capture high-order relations and manage long-term dependencies across tokens through attention mechanisms. This enables transformers to model contextual information effectively. Pretrained language models, such as GPT-3 \citep{brown2020language}, exploit contextual learning to invoke various modules for different tasks, like performing arithmetic upon receiving arithmetic prompts. To further enhance rule comprehension in such models, CoNN-based modules are introduced as a part of Neural Comprehension.

Distinct from common models like BERT \citep{devlin2018bert}, CoNNs leverage a transformer structure and derive their weights from specialized design rather than pretraining. Each Attention layer and Multilayer Perceptron (MLP) layer in a CoNN represents a specific sequence transformation, leading to a neural network module embodying explicit and interpretable operations.

RASP \citep{weiss2021thinking} is a Restricted Access Sequence Processing Language that abstracts the computational model of Transformer-encoder by mapping its essential components, such as attention and feed-forward computation, into simple primitives like select, aggregate, and zipmap. This language enables RASP programs to perform various tasks like creating histograms, sorting, and even logical inference, as demonstrated by \citet{ijcai2020p537}.

Tracr \citep{lindner2023tracr} serves as a compiler that converts human-readable RASP code into weights for a GPT-like transformer architecture with only a decoder module. The Tracr framework uses JAX to transform RASP-defined code into neural network weights. Our neural reasoning framework employs weights generated by Tracr, which are then converted into PyTorch weights to be compatible with the pretrained language model.

Looped Transformers as Programmable Computers \citep{giannou2023looped} introduces a novel transformer framework that simulates basic computing blocks, such as edit operations on input sequences, non-linear functions, function calls, program counters, and conditional branches. This is achieved by reverse engineering attention and hardcoding unique weights into the model, creating a looped structure. The resulting CoNN can emulate a general-purpose computer with just 13 layers of transformers, and even implement backpropagation-based context learning algorithms, showcasing the approach's vast application prospects.

Overall, the potential applications of CoNNs are extensive, given their capacity to perform a wide array of tasks beyond natural language processing. CoNNs offer increased interpretability and transparency through explicitly defined operations, which is vital in fields such as medical diagnosis and legal decision-making. Additionally, CoNNs can lead to more efficient and effective neural network architectures by reducing pretraining requirements and facilitating improved optimization of network parameters.

\subsection{Example}
In this subsection, we briefly describe how computational processes can be represented using transformer code and demonstrate how new CoNN weights can be obtained with the aid of the Tracr compiler.

\subsubsection{\texttt{Parity} CoNN}
In the introduction, we tried to introduce how to perform parity checking on a sequence containing [0 | 1] using a CoNN. Whenever we need to check the sequence, this CoNN can output the completely correct answer.

\begin{table}
\textsc{The tracr code of \texttt{Parity} CoNN}

\vspace{0.5cm}
\begin{lstlisting}
def parity(sop) -> rasp.SOp:
    """Multiply the length of each token."""
    sop = rasp.SequenceMap(lambda x,y: x * y,sop,length).named('map_length') 

    """Add each bit."""
    out = rasp.numerical(rasp.Aggregate(rasp.Select(rasp.indices,rasp.indices,rasp.Comparison.TRUE).named('Select'),rasp.numerical(rasp.Map(lambda x: x, sop).named('map_length')),default=0).named('Aggregate'))  
    
    """Calculate whether the remainder of dividing it by 2 is odd or even."""
    out = rasp.Map(lambda x: 0 if x % 2 == 0 else 1,out).named('Zipmap') 
    
    return out
\end{lstlisting}
\end{table}

\begin{figure}[h]
\begin{center}

  \begin{minipage}{1.17\textwidth}
    \hspace{-0.3cm}
    \includegraphics[width=0.9\textwidth]{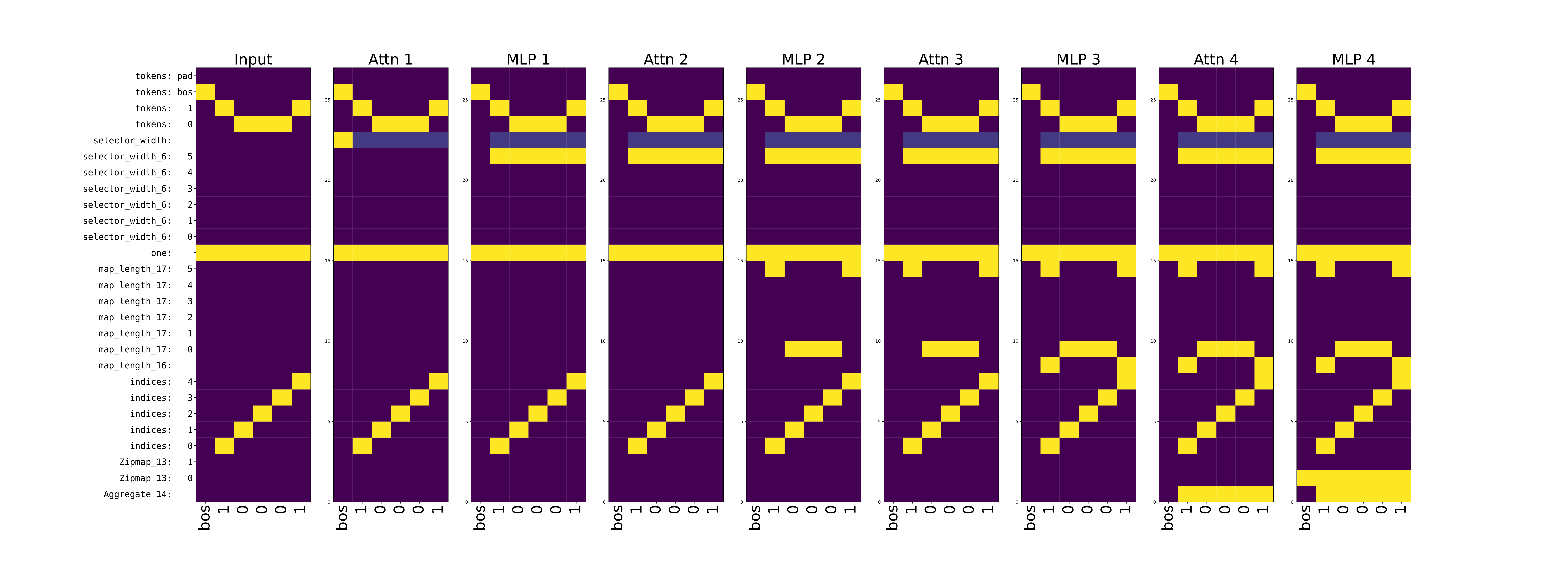}

    \end{minipage}

   \caption{Input the \textit{[1,0,0,0,1]} (target output = 0) for \texttt{Parity} CoNN.}
    \label{figure:7}

    \begin{minipage}{1.17\textwidth}
    \hspace{-0.3cm}
    \includegraphics[width=0.9\textwidth]{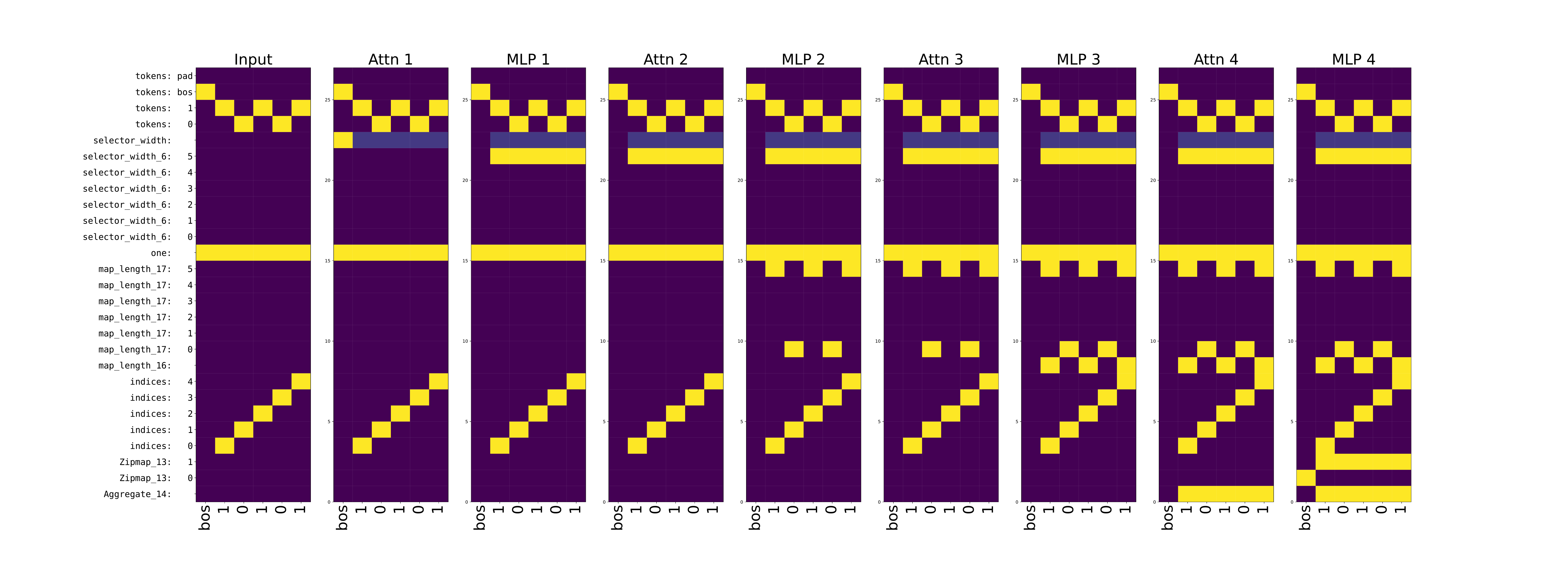}

    \end{minipage}

   \caption{Input the \textit{[1,0,1,0,1]} (target output = 1) for \texttt{Parity} CoNN.}
       \label{figure:8}
  \end{center}
  \end{figure}

Figures \ref{figure:7} and \ref{figure:8} present two distinct input sequences, and illustrate the corresponding hidden state and final output obtained after passing through the internal layers of the \texttt{Parity} CoNN architecture.

\subsubsection{\texttt{Reverse} CoNN}
Figures \ref{figure:9} and \ref{figure:10} show the hidden state and output of Reverse CoNN when inputting text. The embedding of CoNN can be customized, so tokens can be either words like 'hello' or individual letters.

\begin{table}
\textsc{The tracr code of \texttt{Reverse} CoNN}
\vspace{0.8cm}
\begin{lstlisting}
def reverse(sop) -> rasp.SOp:
    """Get the indices from back to front.""" 
    opp_idx = (length - rasp.indices).named("opp_idx")

    """opp_idx - 1, so that the first digit of indices = 0."""
    opp_idx = (opp_idx - 1).named("opp_idx-1")  

    """Use opp_idx to query indices, get the Select."""
    reverse_selector = rasp.Select(rasp.indices, opp_idx,rasp.Comparison.EQ).named("reverse_selector")  

    """Aggregate the reverse_selector and sop"""
    return rasp.Aggregate(reverse_selector, sop).named("reverse")
    

\end{lstlisting}
\end{table}

\begin{figure}[h]
\begin{center}

  \begin{minipage}{1.17\textwidth}
    \hspace{-0.2cm}
    \includegraphics[width=0.85\textwidth]{reverse0.pdf}

    \end{minipage}
\vspace{-0.75cm}
   \caption{Input the \textit{['hello',',','world']} for \texttt{Reverse} CoNN.}
    \label{figure:9}

    \begin{minipage}{1.17\textwidth}
    \hspace{-0.2cm}
    \includegraphics[width=0.85\textwidth]{reverse1.pdf}

    \end{minipage}
\vspace{-0.75cm}
   \caption{Input the \textit{['r','e','v','e','r','s','e']} for \texttt{Reverse} CoNN.}
       \label{figure:10}
       
  \end{center}
  \end{figure}

\subsubsection{\texttt{Addition} CoNN}

Due to the high complexity of the model, we decided to omit the hidden state transformation for the \texttt{Addition} CoNN. However, we have provided code later in the text that will allow for easy implementation of this CoNN. The code includes \texttt{add\_in\_the\_same\_position} and \texttt{add\_carry} functions, which are used to calculate the addition and carry of pairs in the CoNN respectively. We divide the entire operation into two models. For the \texttt{add\_carry} model, we refer to the approach of ALBERT. After the output of the \texttt{add\_in\_the\_same\_position} model, we cyclically use the \texttt{add\_carry} model $\mathrm{L}$ times, where $\mathrm{L}$  is the length of the text, to ensure that all digits can carry. It is important to note that this particular \texttt{Addition} CoNN is only capable of performing addition operations on natural numbers.

\begin{table}
\textsc{The tracr code of \texttt{Addition} CoNN}
\vspace{0.5cm}
\begin{lstlisting}
def split(sop, token, index):
    """Match the position of target token"""
    target_position = rasp.Aggregate(rasp.Select(sop, rasp.Map(lambda x: token, sop), rasp.Comparison.EQ), rasp.indices)  
    
    """If need to match the front position."""
    if index == 0:
        out = rasp.Aggregate(rasp.Select(rasp.indices, rasp.indices - (length - target_position), rasp.Comparison.EQ),
                             sop) # Move the sop on the left side of the token to the far right.
        return rasp.SequenceMap(lambda x, i: x if i == 2 else "_", out, rasp.categorical(
            rasp.SequenceMap(lambda x, i: 2 if x >= i else 0, rasp.indices, length - target_position))) # Use "_" to fill the empty position on the left.

    """If need to match the finally number."""
    else:
        return rasp.SequenceMap(lambda x, i: x if i else "_", sop,
                                rasp.SequenceMap(lambda x, i: 1 if x > i else 0, rasp.indices, target_position)).named(
            f"shift") # Use "_" to fill the empty position on the left.

def atoi(sop):
    """Converts all text to number, and uses 0 for strings of types other than numbers, It may be mixed with 'str' or 'int'.
    """

    return rasp.SequenceMap(lambda x, i: int(x) if x.isdigit() else 0, sop, rasp.indices).named(
        "atoi")

def shift(sop):
    """Get the target indices."""
    idx = (rasp.indices - 1).named("idx-1")


    """Use opp_idx to query indices, get the Select."""
    selector = rasp.Select(idx, rasp.indices,
        rasp.Comparison.EQ).named("shift_selector")  


    """Aggregates the sops and selectors (converted from indexes)."""
    shift = rasp.Aggregate(selector, sop).named("shift")  
    return shift

def add_in_the_same_position(sop):
    x = atoi(split(sop,'+',0)) + atoi(split(sop,'+',1))
    return x

def carry(sop):
    weight = shift(rasp.Map(lambda n:1 if n>9 else 0,sop))
    
    weight = rasp.Aggregate(rasp.Select(rasp.indices,rasp.indices,lambda key,query:key == query),weight,default=0)
    x = rasp.Map(lambda n:n-10 if n>9 else n,sop)
    return x + weight
\end{lstlisting}
\end{table}

\subsubsection{\texttt{Subtraction} CoNN}
The subtraction CoNN is similar to the addition CoNN. First, each digit is subtracted from its corresponding digit, and then it is determined whether to carry over. For ease of experimentation, this subtraction CoNN only supports subtraction of natural numbers where the minuend is greater than the subtrahend.

\begin{table}
\textsc{The tracr code of \texttt{Subtraction} CoNN}
\vspace{0.5cm}
\begin{lstlisting}
def split(sop, token, index):...

def atoi(sop):...

def shift(sop):...

    
def sub_in_the_same_position(sop):
    x = atoi(split(sop,'-',0)) - atoi(split(sop,'-',1))
    return x


def carry(sop):
    weight = shift(rasp.Map(lambda n:1 if n<0 else 0,sop))
    weight = rasp.Aggregate(rasp.Select(rasp.indices,rasp.indices,lambda key,query:key == query),weight,default=0)
    x = rasp.Map(lambda n:n+10 if n<0 else n,sop)
    return x - weight

\end{lstlisting}
\end{table}

\subsection{CoNN model parameters}
\label{appendix:connparameter}
\begin{table*}[htp]
\resizebox{\textwidth}{!}{
\begin{tabular}{lcccccccc}
\midrule \midrule
Model & Layers  &Heads& Vocabulary Size &Window Size& Hidden Size& MLP Hidden Size&\# Parameters &  Compared to GPT-3  \\
\midrule

\texttt{Pariity} & 4 &1& 4 & 40 &132&1959 & 2.2M&$\approx$ 1/100,000 \\
\texttt{Reverse} &4&1&28&40&297&1640&4.3M&$\approx$ 1/50,000 \\
\texttt{Last Letter} &3&1&28&16&103&32&62.6K&$\approx$ 1/3,000,000 \\
\texttt{Copy} &1&1&28&16&69&26&8.8K&$\approx$ 1/20,000,000 \\
{\small \texttt{Add\_in\_the\_same\_position}}&7&1&13&40&535&6422&51.8M&$\approx$1/3000 \\
\texttt{Add\_Carry}&3&1&122&40&130&52&117K&$\approx$1/1,500,000\\
{\small \texttt{Sub\_in\_the\_same\_position}}&7&1&13&40&535&6422&51.8M&$\approx$1/3000\\
\texttt{Sub\_Carry}&3&1&122&40&130&52&117K&$\approx$1/1,500,000\\
\midrule \midrule
\end{tabular}}
\vspace{-0.4cm}
\caption{We reported on a CoNN with a single function, including its actual parameter size and comparison with the parameters of GPT-3.}
\vspace{-0.45cm}
\label{tab:connmodels}
\end{table*}

The parameter sizes of all CoNN models used in this work are listed in Table \ref{tab:connmodels}. It is noteworthy that even for GPT-3, which has parameters that are orders of magnitude larger, it remains challenging to solve symbolic problems. However, with the use of compiled neural networks, only a small number of parameters are needed to achieve Neural Comprehension.

\subsection{Environmental and Human-centric Benefits of Compiled Neural Networks}
\label{appendix:societal}
Compiled Neural Networks (CoNNs) address concerns related to the environmental impact of training large models and the need for human-centric computing. CoNN models can reduce energy consumption and carbon emissions by minimizing extensive pretraining and decreasing parameter size, as seen in Table \ref{tab:connmodels}. This reduction in computational power and energy requirements makes both the training and inference processes more environmentally friendly.
Additionally, Neural Comprehension offers a more interpretable and transparent alternative to conventional deep learning models. CoNN's explicit operation definitions and specialized architecture enable users to comprehend the reasoning behind model decisions, fostering trust and facilitating human-AI collaboration. Increased interpretability also allows for scrutiny of model behavior, promoting the development of fair, accountable, and transparent systems aligned with ethical considerations and human values.'

\section{Experiment for AutoCoNN}
\label{appendix:autoconn}

\subsection{Method}

\begin{wraptable}{r}{0.6\textwidth}
\begin{minipage}{0.6\textwidth}
\centering
\vspace{-0.35cm}

\renewcommand\arraystretch{1.5}
\resizebox{\textwidth}{!}{%
\begin{tabular}{c|lll}

\midrule \midrule
\multicolumn{1}{c}{CoNN Model} &~~Example=1~~~~~~&~~Example=2~~~~~~ &~~Example=5~~~~~~ \\
\hline
 \texttt{Parity} Model&~~5/10&~~10/10&~~10/10
   \\
 \texttt{Reverse} Model&~~10/10&~~10/10&~~10/10
   \\
    \texttt{Last Letter} Model&~~9/10&~~10/10&~~10/10
   \\
    \texttt{Copy} Model&~~10/10&~~10/10&~~10/10
   \\
\midrule \midrule
\end{tabular}}

\caption{For each CoNN model, we selected ten groups of models that were judged to be correct by AutoCoNN. We manually evaluated whether these models were indeed correct. The 'Example=x' means that x Examples were provided in the validation stage.}
\label{tab:conn}
\vspace{-0.2cm}
\end{minipage}
\end{wraptable}

For experts, they may need to spend a lot of time writing code suitable for CoNN, while non-expert users find it hard to obtain or modify CoNN. These issues limit the efficient combination of CoNN and LM, so we utilized the few-shot ability of language models to make the AutoCoNN toolkit \citep{weng2023lmtuner}. In this section, we will show a series of detailed experiments on AutoCoNN to demonstrate this. First, It is the Demo of AutoCoNN code:

\begin{table}[h]
\textsc{Demo of AutoCoNN}
\vspace{0.5cm}
\begin{lstlisting}
from NeuralCom.AutoCoNN import AutoCoNN

INSTRUCT = 'Create an SOp that is the last letter of a word'
VOCAB = ['a','b','c','d','e','f','g']
EXAMPLE = [[['a','b','c'],['c','c','c']],[['b','d'],['d','d']]]

auto = AutoCoNN()
model,tokenizer = auto(instruct=INSTRUCT,vocab=VOCAB,example=EXAMPLE)

\end{lstlisting}
\end{table}

Table \ref{tab:autoconn} shows the efficiency comparison between experts and AutoCoNN. This demonstrates that the AutoCoNN toolkit can generate various CoNNs faster. But we also found that for more difficult ones like Addition and Subtraction, it fails to successfully generate, which becomes one of the limitations of AutoCoNN. On the other hand, we tried providing only "Instruct" or "Example" for AutoCoNN to generate\footnote{In this experiment, the few-shot samples also contained only one of them}, and often "Instruct" can generate CoNN with higher accuracy, while "Example" cannot. This shows that giving explicit operational instructions performs better than directly observing data in AutoCoNN.

\begin{table}[h]
\centering
\renewcommand\arraystretch{1.5}
\resizebox{\textwidth}{!}{%
\begin{tabular}{c|lllll}

\midrule \midrule
\multicolumn{1}{c}{CoNN Model} &Expert's Working Time&Success by AutoCoNN&Can AutoCoNN solve & AutoCoNN (w. Instruct) & AutoCoNN (w. Example)\\
\hline
  \texttt{Parity} Model&1 hours&8/20&\efficacyhigh &7/20 & 3/20\\
\texttt{Reverse} Model&0.5 hour&15/20&\efficacyhigh & 16/20 & 11/20\\
\texttt{Last Letter} Model&0.5 hour&13/20&\efficacyhigh & 12/20 & 10/20\\
\texttt{Copy} Model&0.2 hour&17/20&\efficacyhigh & 17/20 & 15/20\\
\texttt{Addition} Model&48 hours&0/20&\efficacylow & 0/20&0/20\\
\texttt{Subtraction} Model&48 hours&0/20&\efficacylow & 0/20& 0/20\\
\midrule \midrule

\end{tabular}}
\vspace{0.3cm}
\caption{Comparison between AutoCoNN and Expert Built CoNN. The column 'Expert's Working Time' refers to the time required for a trained engineer to write the CoNN code; 'Success by AutoCoNN' refers to the accuracy of 20 results generated by using GPT-3.5 for diverse decoding; 'Can AutoCoNN solve' refers to whether AutoCoNN can identify suitable CoNN code from the 20 results through validation. It is worth noting that in this experiment, we use sampling decoding with temperature=0.7 to generate 20 different CoNNs codes, which we convert to Pytorch versions of CoNNs models. We report the accuracy of the CoNNs codes through manual (expert) evaluation.}
\label{tab:autoconn}
\end{table}

It is difficult for non-expert users to assess the accuracy of the generated code, we automatically utilize the Example information to verify the accuracy of the CoNN model - checking whether the output result of the input sequence is exactly consistent with the Example. The results shown in Table \ref{tab:conn} demonstrate that generally 2 Examples are sufficient to select an accurate CoNN model, which means it is very easy for users to use and demonstrate. However, considering the varying difficulty of different tasks, we still suggest non-expert users provide more Examples to ensure the accuracy of the generated CoNN.

\section{Experimental Settings}

 In this study, we primarily explore the capacity of language models to address symbolic reasoning tasks, concentrating on three areas: symbolic operations, symbolic reasoning, and arithmetic reasoning.

\textbf{Symbolic Operations} \quad Building upon the approaches developed by \citet{anilexploring} and \citet{qian2022limitations}, we examine the following tasks: Parity, Reverse, Addition and Subtraction. These tasks do not require complex text understanding, but only require faithfully implementing symbolic operations and outputting the corresponding results.

\textbf{Symbolic Reasoning} \quad We employ the experimental framework of \citet{weichain} for the two tasks, Last Letter Concatenation and Coin Flip. These tasks require a combination of language understanding and rule comprehension abilities.

\textbf{Arithmetic Reasoning} \quad To evaluate the method's generalization ability from symbolic operations to arithmetic reasoning in addition and subtraction tasks, we use five established arithmetic reasoning datasets: AddSub \citep{MohammadJavadHosseini2014LearningTS}, SingleEq \citep{RikKoncelKedziorski2015ParsingAW}, MultiArith \citep{SubhroRoy2016SolvingGA}, GSM8K \citep{cobbe2021training}, and SVAMP \citep{PatelArkil2021AreNM}. Additionally, we introduce the $\text{AddSub}^{+}$ dataset, containing tasks of varying complexity based on the number of digits involved in arithmetic operations, ranging from 1-digit addition to 20-digit addition/subtraction tasks.

\section{Supplementary Experiment}
\subsection{The effect of training data scale on length generalization of gradient-based models}
\label{appendix:datascale}

\begin{figure}[h]
  \centering

  \subfigure[Parity]{
    \includegraphics[width=0.4\textwidth]{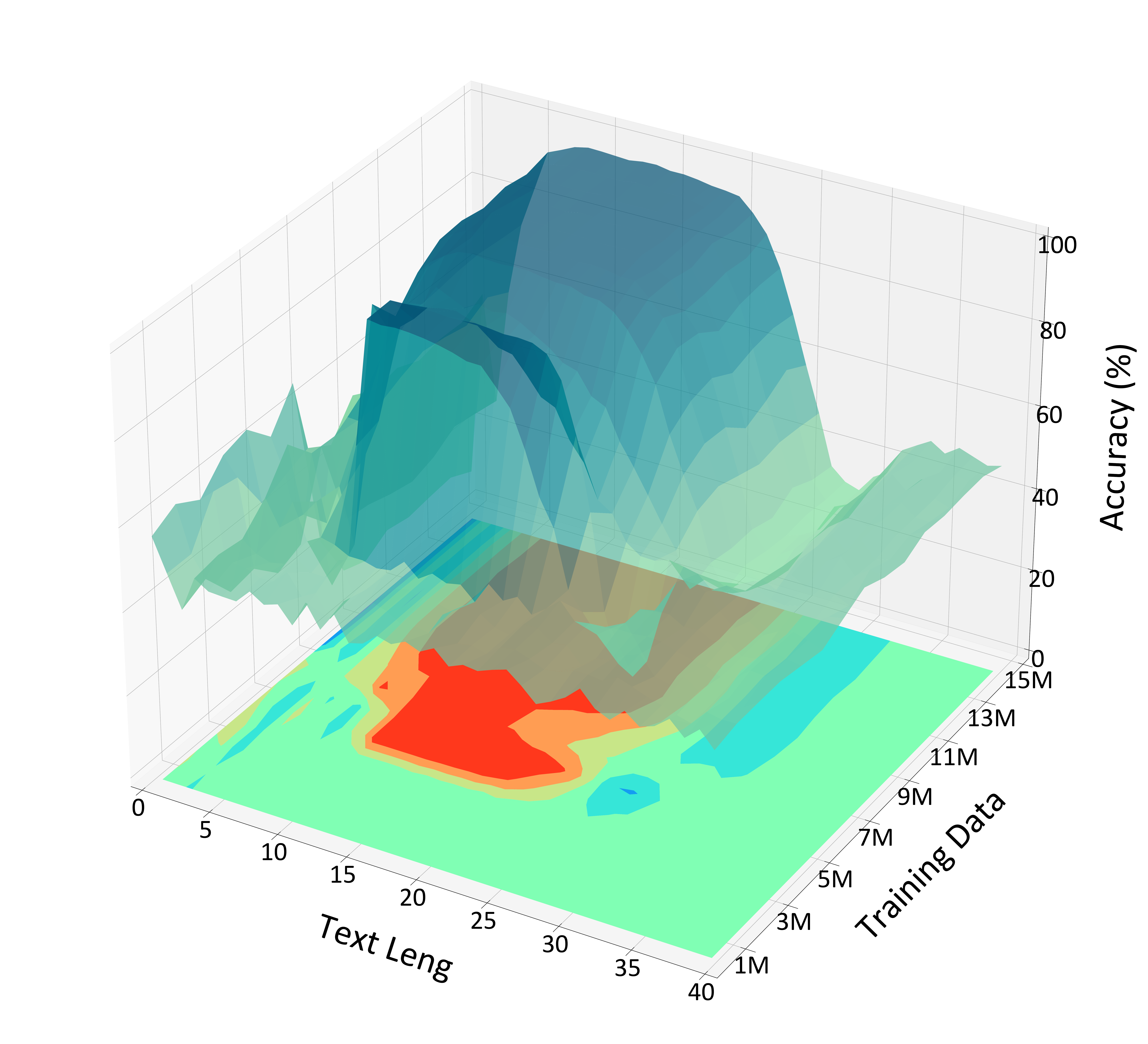}
  }\hspace{7mm}
  \subfigure[Reverse]{
    \includegraphics[width=0.4\textwidth]{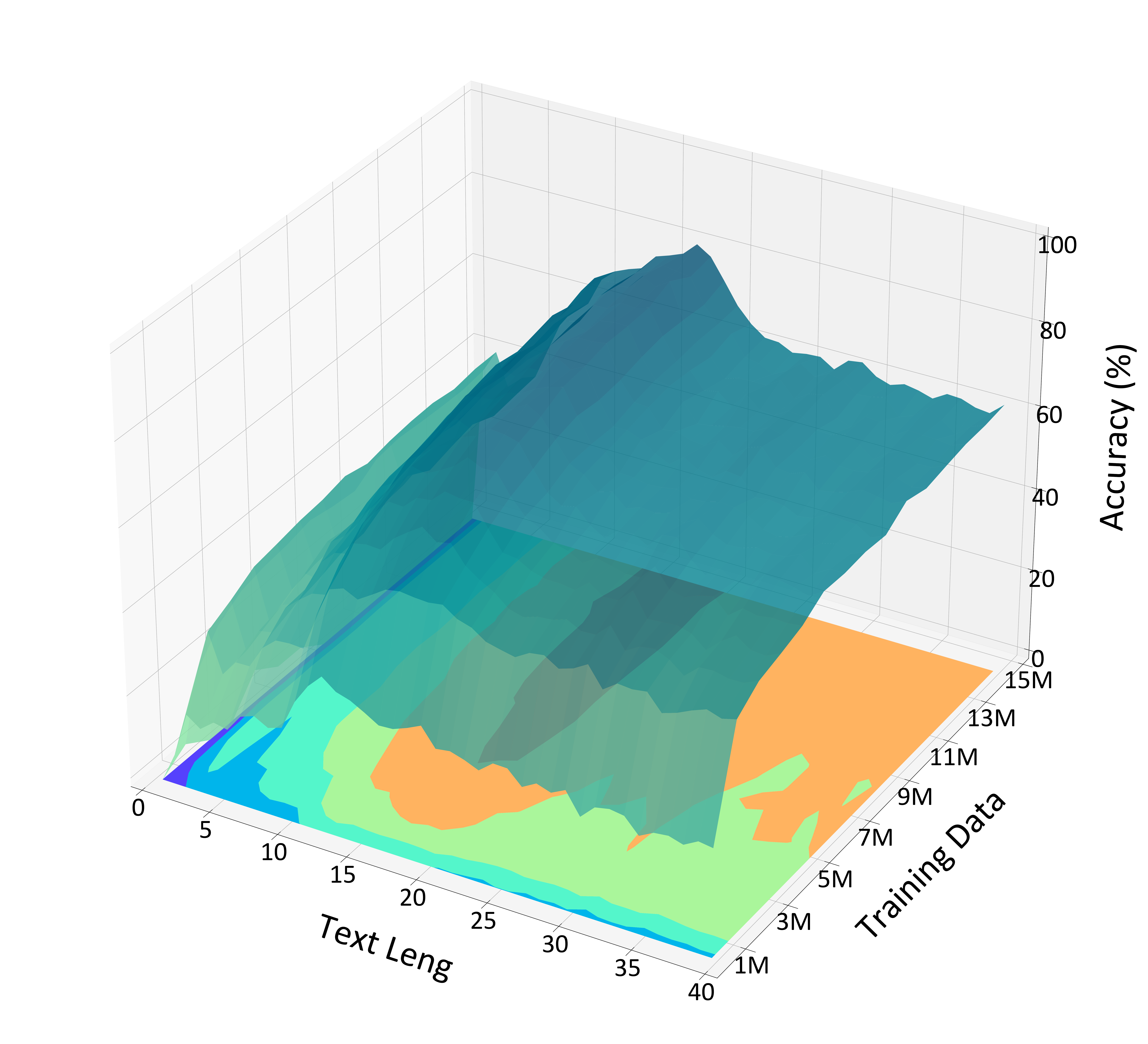}
  }
    \subfigure[Addition]{
    \includegraphics[width=0.4\textwidth]{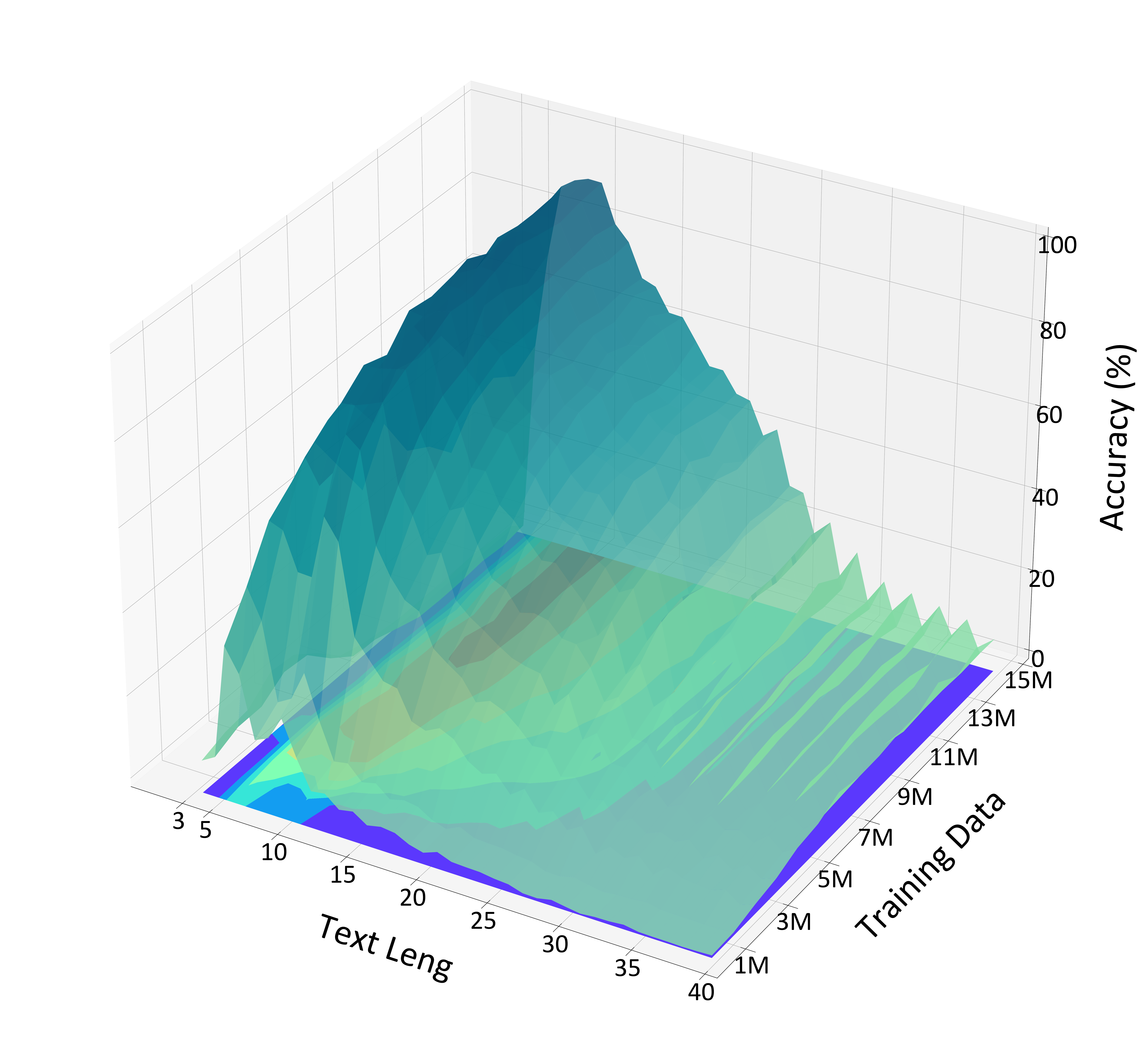}
  }\hspace{7mm}
    \subfigure[Subtraction]{
    \includegraphics[width=0.4\textwidth]{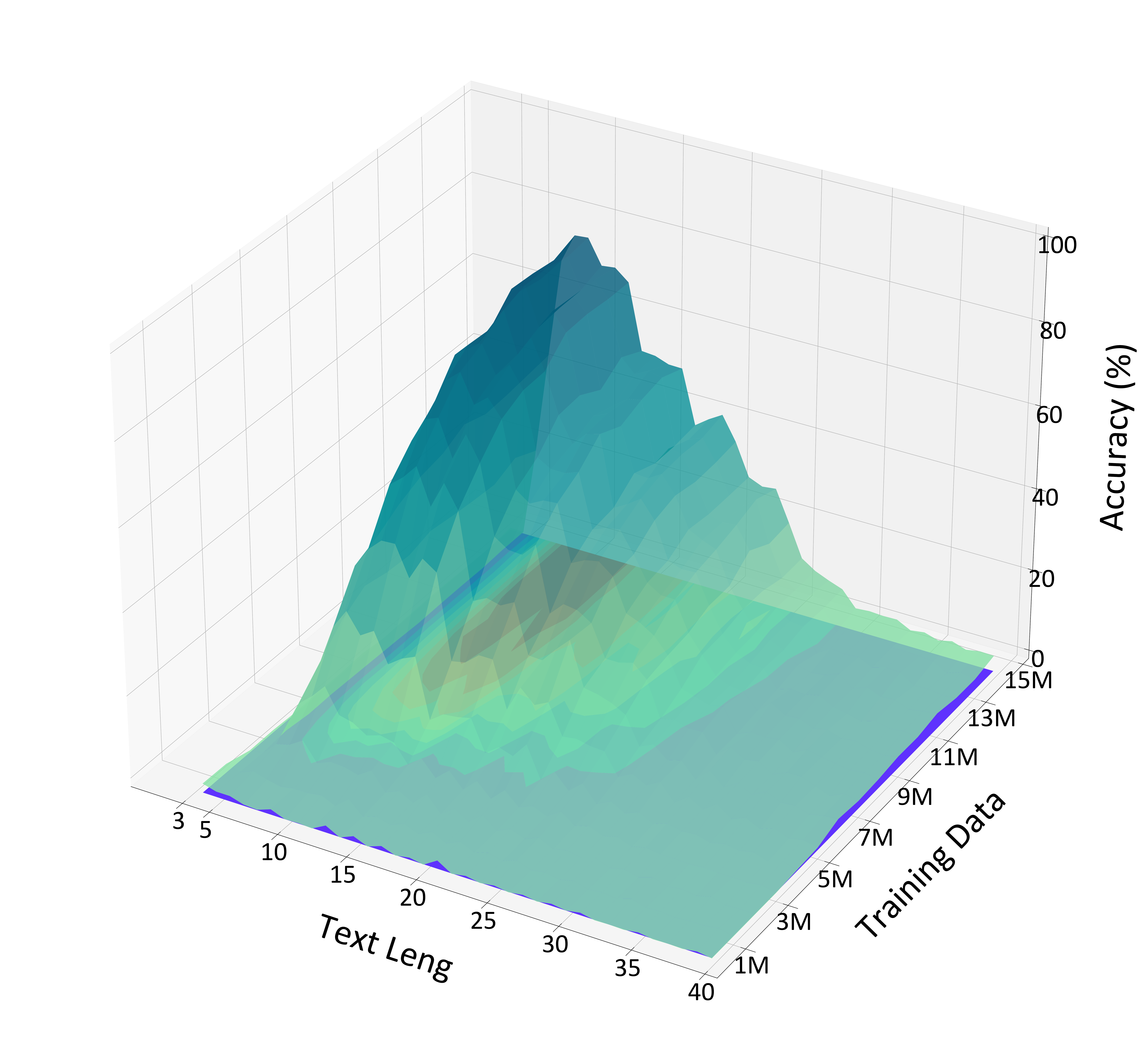}
  }
  \vspace{0.35cm}
  \caption{Length Generalization Performance of Language Models with Different Dataset Sizes.}
    \label{figure:4}
    \vspace{0.35cm}
\end{figure}

To investigate the impact of training data scale on out-of-distribution (OOD) performance, we conducted experiments using the T5-large model with varying amounts of in-distribution training data. The experimental setup closely followed that of \textbf{Main Figure 3}, utilizing numbers with 10 to 20 digits as the training set but varying the number of training examples between 1 million and 15 million. The peak validation set performance for each experiment is reported in Figure \ref{figure:4}.

The results in Figure \ref{figure:4} show that increasing the scale of the In-Dist training data leads to only marginal improvements in OOD performance. This finding is discouraging, suggesting that gradient-based language models face challenges in capturing the true underlying meaning of symbols and their transformation rules based on the data distribution alone.

\subsection{Real-world Arithmetic Reasoning Tasks}
\label{appendix:arithmetic}

\begin{table}[h]

\begin{center}

\renewcommand\arraystretch{1.5}
\resizebox{\textwidth}{!}{%
\begin{tabular}{cc|llllllll}

\midrule \midrule
\multicolumn{2}{c}{Method} &GSM8K~~~~~~~~~~~~&SingleEq~~~~~~~~~~~~ &AddSub~~~~~~~~~~~~&MultiArith~~~~~~~~~~~~&SVAMP~~~~~~~~~~~~&\textbf{Average} \\
  \hline
  \multicolumn{2}{c}{Previous SOTA (Fintune)} &$35^a$/$57^b$&$32.5^c$&$94.9^d$&$60.5^e$&$57.4^f$&- \\
 \multicolumn{2}{c}{GPT-3 Standard}&19.7&86.8&90.9&44.0&69.9&62.26 \\
  \hline
  \multirow{2}{3.1cm}{\centering GPT-3 (175B)\\{\tt{}code-davinci-001}}&CoT&13.84&62.02&57.22&45.85&38.42&43.47   \\
  &CoT + Neural Comprehension&13.95$_{\textcolor[RGB]{219,69,53}{(+0.11)}}$&62.83$_{\textcolor[RGB]{219,69,53}{(+0.81)}}$&60.25$_{\textcolor[RGB]{219,69,53}{(+3.03)}}$&45.85$_{\textcolor[RGB]{40,40,40}{(+0.0)}}$&38.62$_{\textcolor[RGB]{219,69,53}{(+0.2)}}$&44.30$_{\textcolor[RGB]{219,69,53}{(+0.83)}}$ \\
%
  \hline
  \multirow{2}{3.1cm}{\centering GPT-3.5 (175B)\\{\tt{}code-davinci-002}}&CoT&60.20&91.01&82.78&96.13&75.87&81.20   \\
  &CoT + Neural Comprehension&$60.42_{\textcolor[RGB]{219,69,53}{(+0.22)}}$&$91.01_{\textcolor[RGB]{40,40,40}{(+0.0)}}$&$82.78_{\textcolor[RGB]{40,40,40}{(+0.0)}}$&$96.13_{\textcolor[RGB]{40,40,40}{(+0.0)}}$&$76.09_{\textcolor[RGB]{219,69,53}{(+0.22)}}$&$81.29_{\textcolor[RGB]{219,69,53}{(+0.09)}}$ \\
\midrule \midrule
\end{tabular}
}
\end{center}

\caption{Problem solve rate (\%) on arithmetic reasoning datasets. The previous SoTA baselines are obtained from: (a) GPT-3 175B finetuned \citep{cobbe2021training}; (b) GPT-3 175B finetuned plus an additional 175B verifier\citep{cobbe2021training}; (c) \citet{MinghaoHu2019AMM}; (d) \citet{SubhroRoy2016SolvingGA}; (e) \citet{SubhroRoy2016SolvingGA}; (f) \citet{AidaAmini2019MathQATI}; (f) \citet{XinyuPi2022ReasoningLP}}
\label{tab:8}

\vspace{0.5cm}
\end{table}

As model parameters, training calculations, and dataset sizes have increased, language models have gained new capabilities \citep{AarohiSrivastava2022BeyondTI, wei2022emergent}, such as Machine Translation \citep{MIR-2022-09-288,li2024towards}, complex QA \citep{zhu2022reasonchainqa, daull2023complex}, Multimodal QA \citep{MIR-2022-07-224,li2023learning,weng2023visual}, coding \citep{li2022competition, nijkamp2022codegen}, few-shot learning \citep{brown2020language, perez2021true}, medical diagnosis \citep{li2021more, xia2022medconqa}, and chain of thought \citep{weichain, weng2022large}.

In Table \ref{tab:8}, we compared Vanilla CoT with the Neural Comprehension framework for arithmetic reasoning tasks. We integrated the \texttt{Addition} and \texttt{Subtraction} CoNNs with LLMs and observed improved performance across several tasks. This suggests that the proposed Neural Comprehension framework can compensate for the difficulties faced by large-scale language models in computational tasks. Nevertheless, the performance improvement is not as significant due to the choice of specific CoNN models to ensure clarity in our experiments. Designing CoNN models to support more general arithmetic tasks could potentially yield more substantial improvements. In addition, since the Neural Comprehension framework improves the gap between the data distribution learned by the language model during training through gradient descent and the real rules, it can also be combined with some existing logical improvements to language models, including self-consistency \citep{XuezhiWangSelfConsistencyIC}, least-to-most \citep{zhou2022least}, self-improve \citep{huang2022large}, and self-verification \citep{weng2022large}. It can also be combined with some zero-shot methods \citep{TakeshiKojimaLargeLM,ZhuoshengZhang2022AutomaticCO}.

\begin{wraptable}{r}{0.455\textwidth}
\begin{minipage}{0.455\textwidth}
\centering

\renewcommand\arraystretch{1.5}
\resizebox{\textwidth}{!}{%
\begin{tabular}{c|lll}

\midrule \midrule
\multicolumn{1}{c}{Method} &~~T5-small~~~~~~&~~T5-base~~~~~~ &~~T5-large~~~~~~ \\
\hline
  Origin&~~1.74&~~1.52&~~3.87
   \\

  Neural Comprehension& ~~1.82&~~1.59&~~4.02  \\
  \hline

Ours Improve&+0.08&+0.07&+0.15\\
\midrule \midrule
\end{tabular}}

\caption{The test set problem-solving rate (\%) of the T5 model on the GSM8K dataset.}
\label{tab:problemt5}
\vspace{-0.2cm}
\end{minipage}
\end{wraptable}

To further evaluate the effectiveness of the Neural Comprehension framework, Table \ref{tab:problemt5} presents the results of fine-tuning T5 models with \texttt{Addition} and \texttt{Subtraction} CoNN on the GSM8K training dataset. The comparison of three different-sized models reveals that the framework can model deterministic rules defined by humans, thus avoiding the uncertainty associated with gradient descent learning from data distribution.

\subsection{The Efficiency of Neural Comprehension}

\begin{table}[h]

\renewcommand\arraystretch{1.5}
\resizebox{\textwidth}{!}{%
\begin{tabular}{ccc|rr|rr|rr}

  \midrule  \midrule
  &&&\multicolumn{2}{c|}{Vanilla}&\multicolumn{2}{c|}{Neural Comprehension}&\multicolumn{2}{c}{$\delta_{\textbf{Time}}$} \\
\multicolumn{1}{c}{Model} & Params &Task &GPU &CPU &GPU & CPU&GPU & CPU    \\

\midrule
  T5-small&60M&Coin Flip&5.280s&5.720s&5.431s&5.872s&0.151s (2.86\%)&0.152s (2.66\%)\\
T5-base&220M&Coin Flip&7.865s&13.767s&8.010s&13.939s&0.145s (1.84\%)&0.172s (1.25\%)\\
T5-large&770M&Coin Flip&14.055s&32.953s&14.194s&33.120s&0.139s (0.99\%)&0.167s (0.51\%)\\
  \midrule
    T5-small&60M&Last Letter Concatenation&16.233s&28.309s&16.744s&28.720s&0.511s (3.15\%)&0.411s (1.45\%)\\
T5-base&220M&Last Letter Concatenation&28.912s&55.660s&29.426s&56.087s&0.514s (1.78\%)&0.427s (0.77\%)\\
T5-large&770M&Last Letter Concatenation&49.584s&103.739s&50.066s&104.134s&0.482s (0.97\%) & 0.395s (0.38\%)\\
    \midrule  \midrule
\end{tabular}}
\caption{In Neural Comprehension framework, the inference latency comparison of the T5 model.}
\label{tab:inference}
\end{table}

To evaluate the efficacy of Neural Comprehension, we conducted further experiments comparing the inference latency of both the Vanilla and Neural Comprehension frameworks on an equal number of sequences and equal sequence lengths using GPU and CPU configurations. We employed a batch size of 1 and assessed the inference latency of Neural Comprehension in conjunction with various T5 model sizes across two symbolic inference tasks to ascertain efficiency. The full results are detailed in Table \ref{tab:inference}.

Our findings reveal that implementing Neural Comprehension increases computational requirements, primarily attributed to the supplementary parameter volume and computational demands introduced by CoNNs. However, as the scale of pretrained language models expands, the proportion of $\delta_{\textbf{Time}}$ within the Neural Comprehension framework progressively diminishes, particularly for larger language models.

\section{Implementation and Details}
\label{appendix:Implementation}

\begin{table*}[h]
\resizebox{\textwidth}{!}{
\begin{tabular}{lccccccc}
\midrule \midrule
Model & Model Creator & Modality&Version & \# Parameters & Tokenizer & Window Size & Access \\
\midrule

T5-small & Google & Text & T5.1.0&60M & T5 & 512&Open \\
T5-base & Google & Text & T5.1.0&220M & T5 & 512&Open \\
T5-large & Google & Text & T5.1.0&770M & T5 & 512&Open \\
\midrule

GLM-130B & Tsinghua University & Text &GLM-130B& 130B & ICE & 2048 & open \\
GPT-3 & OpenAI & Text,Code &code-davinci-001& 175B*  & GPT-2 & 2048 & limited \\
GP-3.5 & OpenAI & Text,Code,Instruct &code-davinci-002& 175B*  & GPT-2 & 8096 & limited  \\
GPT-4 & OpenAI &Text,Code,Instruct,$\dots$ &gpt-4& 175B0*  & GPT-2 & 8000 & limited  \\

\midrule \midrule
\end{tabular}}
\vspace{-0.3cm}
\caption{Models. Description of the models evaluated in this effort: provenance for this information is provided in models. $*$ indicates that we believe the associated OpenAI models are this size, but this has not been explicitly confirmed to our knowledge.}
\vspace{-0.3cm}
\label{tab:models}
\end{table*}

In this section, we provide a detailed description of the experimental setup from a model and dataset perspective, ensuring repeatability of our experiments.
\subsection{Model}
\label{appendix:model}
Our experiments primarily involve the T5, GPT, and GLM-130B families of models. Neural Comprehension framework supports seamless integration with language models having decoder structures, regardless of the scale of the language model. We fine-tune the T5 models, while the larger models with over 10 billion parameters are used for few-shot In-context learning. Table \ref{tab:models} presents a comprehensive list of all models used in our experiments\footnote{The specific configurations of the GPT series of models can be found at \url{https://platform.openai.com/docs/model-index-for-researchers/model-index-for-researchers}}.  
\subsubsection{Fine-tuning}

For the T5 models, we employ the standard fine-tuning approach using the pretrained models as a starting point. We follow the pre-processing steps in the T5 original paper, which involves set the input text max length to 150 and using the tokenizer to process the data. We use a batch size of 64 for all models and the Adafactor optimizer \citep{shazeer2018adafactor} with a learning rate of $1\times10^{-4}$. The models are trained for a maximum of 20 epochs. We use a cosine learning rate schedule with a warm-up phase comprising 5\% of the total number of training steps. We employ a dropout rate of 0.1 during training to mitigate overfitting. Our experiments utilize the PyTorch framework \citep{NEURIPS2019_bdbca288} for training and inference. Table 5 displays the parameter settings for the T5 models during training, which is conducted on four NVIDIA A6000 GPUs with 48GB of memory each.

\begin{wraptable}{l}{0.49\textwidth}
\vspace{-0.6cm}
\begin{center}
\begin{tabular}{lr}

\midrule \midrule
\textbf{Training Setting} & \textbf{Configuration} \\\bottomrule
optimizer&Adafactor \\
base learning rate&$1\times10^{-4}$\\
weight decay&$2\times10^{-5}$\\
decay rate&-0.8\\
optimizer eps&{\small ($1\times10^{-30}$,$2\times10^{-3}$)}\\
batch size&64\\
training epochs&20\\
gradient clip&1.0\\
\midrule \midrule
\end{tabular}
\end{center}
\vspace{-0.4cm}
    \caption{Training Setting}
\label{tab:t5model}
\vspace{-0.4cm}
\end{wraptable}

In Table \ref{tab:t5model}, we list the hyperparameters used to train the T5 model. We carefully selected these parameters to ensure that the within-distribution validation accuracy roughly converged. We report all peak validation set results, and in every experiment we ran, we found that within-distribution validation accuracy monotonically increased during training iterations (until it approached 100\%), and we never observed overfitting. This may be due to the regular nature of the tasks we considered in the paper. We followed the \citet{anilexploring}'s setup and did not use OOD performance in model selection, as this would constitute "peaking at the test conditions". Regarding the number of training iterations, we also tried training the T5 model with more iterations in the addition task, but this did not lead to substantial differences in OOD performance (i.e., it was still equally poor).

\subsubsection{Few-shot In-Context Learning}
For the few-shot context learning on GPT and GLM-130B models, we employ an approach inspired by the recent work on GPT-3 \citep{brown2020language}. In this methodology, the models are provided a context consisting of a few examples of the input-output pairs in natural language format. Following this, the models are expected to generalize and perform well on the task without any explicit fine-tuning. We carefully design the context to include diverse examples that represent the range of input types and required model reasoning. Importantly, we limit the context length to be within the maximum token limits of the models. For instance, GPT-3 has a token limit of 2048. Due to limited access to the GPT family of models, we utilize the official API for these experiments \footnote{OpenAI's API:\url{https://openai.com/api/}}. .For the GLM-130B, we employ the FasterTransformer framework to set up local inference with INT4 on eight NVIDIA GeForce RTX 3090 GPUs with 24GB of memory each.

To compare with CoT and PAL which experimented on GPT-3 series models, we simulated Neural Comprehension (NC) within the constraints of the API access. We treated the API's output as if it were part of the Neural Comprehension structure. This involved a simulated gating mechanism, where the output from the API was corrected using CoNNs form left to right, and then the adjusted response (Truncate the text after the altered text.) was changed into the API' input for continued generation. This simulation was to ensure that the benefits of NC could be compared fairly with the existing results from PAL.
\subsection{Tasks and Dataset}
\label{appendix:td}
In this paper, all data sets related to length generalization consist of independent data sets with the same number of digits but different lengths, and each digit in the test set is unique. Therefore, there may be slight fluctuations between data sets of different lengths, but the overall trend is generally clear. To further illustrate the differences between data sets of different lengths, the following examples are provided:
\begin{equation*}
	\begin{aligned}
			\small
		{\textbf{Parity:}} &\quad\quad \underbrace{1 1 0}_{\text {Length = 3}} = 0 \quad\quad\quad \underbrace{1 0 1 1 0 0}_{\text {Length = 6}} = 1 \quad\quad\quad \underbrace{0 1 0 0 0 1 1 1 0 1 0 1}_{\text {Length = 12}} = 0
		\\ 
		\vspace{1cm}
			\small
		{\textbf{Reverse:}} &\quad\quad \underbrace{\text{abc}}_{\text {Length = 3}} = {\text{cba}} \quad\quad \underbrace{\text{figure}}_{\text {Length = 6}} = {\text{erugif}} \quad\quad \underbrace{\text{accomplished}}_{\text {Length = 12}} = {\text{dehsilpmocca}}
		\vspace{1cm}
		\\
			\small
		{\textbf{Addition:}} &\quad\quad \underbrace{1+2}_{\text {Length = 3}} = 3 \quad\quad\quad \underbrace{18+245}_{\text {Length = 6}} = 263 \quad\quad\quad \underbrace{48864+964315}_{\text {Length = 12}} = 1013179
		\label{eq:24}
		\vspace{1cm}
		\\
			\small
		{\textbf{Arithmetic Reasoning:}} &\quad\quad 
		{\text{Joan found }}\underbrace{6546634574688499}_{\text {Length = 16}}{\text {seashells and Jessica found}}  \\ &\underbrace{3855196602063621}_{\text {Length = 16}} {\text {seashells on the beach. How many seashells did they find}} \\& \text{together ?}
		\end{aligned}
		\end{equation*}
\subsubsection{Data Generation Details}
\textbf{Synthetic Parity Dataset:} We sample instances of lengths 1-40 from a uniform Bernoulli distribution. We first uniformly sample the number of ones, and then randomly shuffle the positions of each one within a fixed-length bitstring. For the experiments in \textbf{Figure} \ref{fig:sybolic}, we train T5 on lengths 10-20, with 99000 training samples per bit. For all methods, we test each bit using 1000 samples.

\begin{table}
\textsc{Parity Dataset}
\vspace{0.4cm}
\begin{lstlisting}
def generate_parity_data(n):
    data = []
    for _ in range(100000):
        input_str = ''.join(str(random.randint(0, 1)) for _ in range(n))
        label = sum(int(x) for x in input_str) % 2
        data.append({'input': input_str, 'label': label})
    return data

parity_data = {}
for n in range(1, 41):
    parity_data[n] = generate_parity_data(n)
\end{lstlisting}
\end{table}
\vspace{0.5cm}

\textbf{Synthetic Reverse Dataset:} We selected a dataset of instances with lengths ranging from 1 to 40. For the experiments in \textbf{Figure} \ref{fig:sybolic}, the training set consists of 99000 samples each for lengths 10-20. Each input is a randomly generated word string of specified length, where the letters are selected uniformly at random from a set of 26 letters using a Bernoulli distribution (without necessarily having any actual meaning), and all letters are converted to lowercase. The test set for lengths 1-40 contains at least 1000 test samples.

\begin{table}
\textsc{Reverse Dataset}
\vspace{0.4cm}
\begin{lstlisting}
reverse_data = {}
for n in range(1, 41):
    reverse_data[n] = []
    for _ in range(100000):
        word = ''.join(random.choice(string.ascii_lowercase) for _ in range(n))
        reverse_data[n].append({'input':word,'label':''.join(list(reversed(word)))})
\end{lstlisting}
\end{table}
\vspace{0.5cm}

\textbf{Synthetic Addition and subtraction Dataset:} Addition and subtraction are fundamental arithmetic operations that are commonly taught in primary school. To generate the dataset, we takes as input the number of digits $n$ and returns a list of 100000 examples. The function first calculates the remainder $k$ when $(n-1)$ is divided by $2$, and then divides $(n-1)$ by $2$ if n is greater than $2$, else it sets n to itself. This ensures that the length of the first number is either equal to or one digit longer than the length of the second number. The function then generates $100000$ examples using randomly generated numbers. Specifically, it generates two numbers a and b where a is a random integer between $10^{(n+k-1)}$ and $10^{(n+k)}-1$, and b is a random integer between $10^{(n-1)}$ and $10^{n}-1$. It then appends each example to the list data in the form of a dictionary with the input as the string "a+b" and the label as the sum a+b. 

\begin{table}
\textsc{Addition Dataset}
\vspace{0.4cm}
\begin{lstlisting}
# Generate two random n-digit numbers and their sum
def generate_additive_example(n):
    data = []
    k = (n - 1) % 2
    n = (n - 1) // 2 if n > 2 else n

    for _ in range(100000):
        a = random.randint(10**(n+k-1), 10**(n+k) - 1)
        b = random.randint(10**(n-1), 10**n - 1)
        data.append({'input': str(a)+'+'+str(b), 'label': a + b})
    return data

# Generate an additive data set for digits ranging from 3 to 40
additive_data = {}
for n in range(3, 41):
    additive_data[str(n)] = generate_additive_example(n)
\end{lstlisting}
\end{table}
\vspace{0.5cm}

For the experiments in Figure \textbf{Figure} \ref{first}, we provided 99000 training data examples for addition with numbers ranging from 3 to 10 digits in length for the T5 model. For the GPT-3.5 and GPT-4 models, we provided 8 few-shot samples within 10 digits. We evaluated the performance of all three models on numbers ranging from 3 to 30 digits in length, with 1000 test samples per digit. On the other hand, for the experiments in Figure \textbf{Figure} \ref{fig:sybolic}, we provided 99000 training data examples for addition with numbers ranging from 10 to 20 digits in length for the T5 model. For the GPT-3.5 and GPT-4 models, we provided 8 few-shot samples within the range of 10 to 20 digits. We evaluated the performance of all three models on numbers ranging from 3 to 30 digits in length, with 1000 test samples per digit.

For subtraction, we use a similar approach.

\begin{table}
\textsc{Subtraction Dataset}
\vspace{0.4cm}
\begin{lstlisting}
# Generate two random n-digit numbers and their minus
def generate_minus_example(n):
    data = []
    k = (n - 1) % 2
    n = (n - 1) // 2 if n > 2 else n
    for _ in range(100000):
        a = random.randint(10**(n+k-1), 10**(n+k) - 1)
        b = random.randint(10**(n-1), 10**n - 1)
        if a > b:
            data.append({'input': str(a)+'-'+str(b), 'label': a - b})
        else:
            data.append({'input': str(b)+'-'+str(a), 'label': b - a})
    return data

# Generate an subtraction data set for digits ranging from 3 to 40
minus_data = {}
for n in range(3, 41):
    minus_data[str(n)] = generate_minus_example(n)
\end{lstlisting}
\end{table}
\vspace{0.5cm}

\subsubsection{Symbolic Reasoning Dataset} 
\textbf{Coin Flip:} We followed \citet{weichain}'s setup and randomly concatenated first and last names from the top 1000 names in the census data (\url{https://namecensus.com/}) to create the \textsc{<NAME>} token. In our work, flipping a coin corresponds to 1 and not flipping a coin corresponds to 0. To make the inputs as close to English as possible without using too many symbols, we used the sentence models "\textsc{<NAME>} flips the coin." and "\textsc{<NAME>} does not flip the coin." to represent whether the coin was flipped or not. This task is similar to the parity task, but requires further semantic understanding. We constructed a training set of 1500 samples, with 500 samples for each of 2-4 coin flips. For the test set, we selected 100 non-overlapping samples for each of 2-4 coin flips, and evaluated the model every 5 steps.

\begin{table}
\textsc{Coin Filp Dataset}
\vspace{0.4cm}
\begin{lstlisting}
dataset = []
for i in range(500):
    # randomly choose two names from the name_list
    for o in range(2,5):
        sentence = 'A coin is heads up.'
        label = []
        for time in range(o):
            name = random.sample(names, k=1)[0]

            # randomly choose whether to flip the coin or not
            flip = random.choice([True, False])

            # generate the statement and label based on whether the coin was flipped or not
            if flip:
                sentence += f" {name.capitalize()} flips the coin."
                label.append(1)
            else:
                sentence += f" {name.capitalize()} does not flip the coin."
                label.append(0)
        sentence += ' Is the coin still heads up?'

        dataset.append({'question':sentence, 'answer':{0:'yes',1:'no'}[sum(label)%2]})
\end{lstlisting}
\end{table}
\vspace{0.5cm}

\textbf{Last Letter Concatenation:} We followed \citet{weichain}'s setup and randomly concatenated first and last names from the top 1000 names in the census data to create the \textsc{<NAME>} token. This task requires the model to connect the last letter of each word in a concatenated name. This task requires Neural Comprehension of rules in two aspects. First, it requires the model to correctly identify the last letter of each word. Second, it requires the model to concatenate all the last letters of the words together. We concatenated 2-5 first or last names, and constructed a training set of 1500 samples, with 500 samples for each name length of 2-4. For the test set, we selected 100 non-overlapping samples for each name length of 2-4, and evaluated the model every 5 steps.

\subsubsection{Arithmetical Reasoning Dataset} 
\begin{table}[h]

\begin{center}
\begin{tabular}{lcccr}
\midrule \midrule
\textbf{Dataset}&\textbf{Number of samples}&\textbf{Average words}&\textbf{Answer Format}&\textbf{Lience} \\ \hline
GSM8K&1319&46.9&Number&MIT License \\
SingleEq&508&27.4&Number&MIT License \\
AddSub&395&31.5&Number&Unspecified \\
MultiArith&600&31.8&Number&Unspecified\\
SVAMP&1000&31.8&Number&MIT License\\
\midrule \midrule

\end{tabular}
\end{center}
\caption{Arithmetical Reasoning Dataset Description.}
\label{table:5}
\end{table}
In Table \ref{table:5}, we summarize the information of all arithmetic reasoning datasets used in this work. We provide the links to access these datasets:
\begin{itemize}
    \item \textbf{GSM8K}: \url{https://github.com/openai/grade-school-math}
    \item \textbf{SingleEq}: \url{https://gitlab.cs.washington.edu/ALGES/TACL2015}
    \item \textbf{AddSub}: \url{https://www.cs.washington.edu/nlp/arithmetic}
    \item \textbf{MultiArith}: \url{http://cogcomp.cs.illinois.edu/page/resource\_view/98} 
    \item \textbf{SVAMP}: \url{https://github.com/arkilpatel/SVAMP}
\end{itemize}

\definecolor{ryellow}{HTML}{F6F1B9}
\definecolor{rbleu}{HTML}{DEEBF7}
\definecolor{rgray}{HTML}{F0F0F0}

\section{Some examples of Neural Comprehension}
\label{appendix:example}
In this section, we will use \colorbox{rgray}{gray font} to represent the task input, \colorbox{ryellow}{yellow font} to represent the neural network output during training, and \colorbox{rbleu}{blue background} to represent the neural network output during generated.

\subsection{Synthetic Symbolic}

\begin{table}[H]
    \centering
    \small

    \begin{tabular}{p{\linewidth}}
        \bottomrule \bottomrule
        \vspace{-2mm}
\colorbox{rgray}{Q: 1011001010 A: 1}\\
\colorbox{rgray}{Q: 01111011000 A: 0}\\
\colorbox{rgray}{Q: 1010011001110 A: 1}\\
\colorbox{rgray}{Q: 10000001001001 A: 0}\\
\colorbox{rgray}{Q: 110100011110001 A: 0}\\
\colorbox{rgray}{Q: 1110011001010110 A: 1}\\
\colorbox{rgray}{Q: 1100000111011000101 A: 1}\\
\colorbox{rgray}{Q: 01100000110110010001 A: 0}\\
--------{\small (LLM's few-shot prompt)}--------\\
\colorbox{rgray}{Q: 011110001010101101011 A: }\colorbox{rbleu}{0}\\
        \bottomrule \bottomrule
    \end{tabular}
        \caption{
    The example of Parity
    }
    \label{english:prompt}
\end{table}

\begin{table}[H]
    \centering
    \small

    \begin{tabular}{p{\linewidth}}
        \bottomrule \bottomrule
        \vspace{-2mm}
\colorbox{rgray}{Q: neofascism A: msicsafoen}\\
\colorbox{rgray}{Q: betaquinine A: eniniuqateb}\\
\colorbox{rgray}{Q: corediastasis A: sisatsaideroc}\\
\colorbox{rgray}{Q: ferroelectronic A: cinortceleorref}\\
\colorbox{rgray}{Q: cryoprecipitation A: noitatipicerpoyrc}\\
\colorbox{rgray}{Q: cryofibrinogenemia A: aimenegonirbifoyrc}\\
\colorbox{rgray}{Q: chemocarcinogenesis A: sisenegonicracomehc}\\
\colorbox{rgray}{Q: ponjpcdqjuuhiviojmby A: ybmjoivihuujqdcpjnop}\\
----------{\small (LLM's few-shot prompt)}----------\\
\colorbox{rgray}{Q: helloworldhellochina A: }\colorbox{rbleu}{anihcollehdlrowolleh}\\
        \bottomrule \bottomrule
    \end{tabular}
        \caption{
    The example of Reverse
    }
    \label{english:prompt}
\end{table}

\begin{table}[H]
    \centering
    \small

    \begin{tabular}{p{\linewidth}}
        \bottomrule \bottomrule
        \vspace{-2mm}
\colorbox{rgray}{Q: 82637+3058 A: 85695}\\
\colorbox{rgray}{Q: 58020+96632 A: 154652}\\
\colorbox{rgray}{Q: 717471+58704 A: 776175}\\
\colorbox{rgray}{Q: 298309+702858 A: 1001167}\\
\colorbox{rgray}{Q: 1061462+2623780 A: 3685242}\\
\colorbox{rgray}{Q: 58720970+61609034 A: 120330004}\\
\colorbox{rgray}{Q: 364920479+78861480 A: 443781959}\\
\colorbox{rgray}{Q: 6050330002+211065324 A: 6261395326}\\
----------{\small (LLM's few-shot prompt)}----------\\
\colorbox{rgray}{Q: 20021012+20021004 A: }\colorbox{rbleu}{40042016}\\
        \bottomrule \bottomrule
    \end{tabular}
        \caption{
    The example of Addition
    }
    \label{english:prompt}
\end{table}

\begin{table}[H]
    \centering
    \small

    \begin{tabular}{p{\linewidth}}
        \bottomrule \bottomrule
        \vspace{-2mm}
\colorbox{rgray}{Q: 82637-3058 A: 79579}\\
\colorbox{rgray}{Q: 96632-58020 A: 38612}\\
\colorbox{rgray}{Q: 717471-58704 A: 658767}\\
\colorbox{rgray}{Q: 702858-298309 A: 404549}\\
\colorbox{rgray}{Q: 2623780-1061462 A: 1562318}\\
\colorbox{rgray}{Q: 68720970-61609034 A: 7111936}\\
\colorbox{rgray}{Q: 364920479-78861480 A: 286058999}\\
\colorbox{rgray}{Q: 6050330002-211065324 A: 393967676}\\
----------{\small (LLM's few-shot prompt)}----------\\
\colorbox{rgray}{Q: 20021012-20021004 A: }\colorbox{rbleu}{8}\\
        \bottomrule \bottomrule
    \end{tabular}
        \caption{
    The example of Subtraction
    }
    \label{english:prompt}
\end{table}

\subsection{Symbolic Reasoning}

\begin{table}[H]
    \centering
    \small

    \begin{tabular}{p{\linewidth}}
        \bottomrule \bottomrule
        \vspace{-2mm}
\colorbox{rgray}{A coin is heads up. Devin flips the coin. Maxwell does not flip the coin. }\\
\colorbox{rgray}{James flips the coin. Kenneth flips the coin. Is the coin still heads up?}\\
\colorbox{ryellow}{1 0 1 1 ->}\colorbox{rbleu}{1}\\
        \bottomrule
        \colorbox{rgray}{A coin is heads up. Ira flips the coin. Danny does not flip the coin. }\\
\colorbox{rgray}{Horace flips the coin. Is the coin still heads up? }\\
\colorbox{ryellow}{1 0 1 ->}\colorbox{rbleu}{0}\\
       \bottomrule \bottomrule
    \end{tabular}
        \caption{
    The example of Coin Flip
    }
    \label{english:prompt}
\end{table}

\begin{table}[H]
    \centering
    \small

    \begin{tabular}{p{\linewidth}}
        \bottomrule \bottomrule
        \vspace{-2mm}
\colorbox{rgray}{Take the last letters of the words in Elias Earnest Milton and concatenate them.}\\
\colorbox{ryellow}{The last letter of Elias ->}\colorbox{rbleu}{s}\colorbox{ryellow}{The last letter of Earnest ->}\colorbox{rbleu}{t}\\\colorbox{ryellow}{The last letter of Milton ->}\colorbox{rbleu}{n}\colorbox{ryellow}{The answer is}\colorbox{rbleu}{stn}\\
        \bottomrule
\colorbox{rgray}{Take the last letters of the words in Randolph Weldon Olin Robbie\" and concatenate them.}\\
\colorbox{ryellow}{The last letter of Randolph ->}\colorbox{rbleu}{h}\colorbox{ryellow}{The last letter of Weldon ->}\colorbox{rbleu}{n}\\\colorbox{ryellow}{The last letter of Olin ->}\colorbox{rbleu}{n}\colorbox{ryellow}{The last letter of Robbie ->}\colorbox{rbleu}{e}\colorbox{ryellow}{The answer is}\colorbox{rbleu}{hnne}\\
       \bottomrule \bottomrule
    \end{tabular}
        \caption{
    The example of Last Letter Concatenation
    }
    \label{english:prompt}
\end{table}

\subsection{Arithmetical Resoning}

\begin{table}[H]
    \centering
    \small

    \begin{tabular}{p{\linewidth}}
        \bottomrule \bottomrule
        \vspace{-2mm}
\colorbox{rgray}{Joan found 65466345746884996 seashells and Jessica found 38551966020636213}\\ \colorbox{rgray}{seashells on the beach . How many seashells did they find together ?}\\
\colorbox{ryellow}{Joan started with 65466345746884996 seashells. She gave some to Sam. So:}\\\colorbox{ryellow}{ 65466345746884996 - 38551966020636213 =}\colorbox{rbleu}{2691437972624878}\\\colorbox{ryellow}{The answer is 2691437972624878}\\
       \bottomrule \bottomrule
    \end{tabular}
        \caption{
    The example of Arithmetirc Reasoning
    }
    \label{english:prompt}
\end{table}

\end{document}